\documentclass{article}

\usepackage[dblblindworkshop, final]{neurips_2026}
\workshoptitle{}
\makeatletter
\renewcommand{\@noticestring}{%
  \@neuripsordinal\ Conference on Neural Information Processing Systems
  (NeurIPS \@neuripsyear). Workshop.%
}
\makeatother

\usepackage[utf8]{inputenc} % allow utf-8 input
\usepackage[T1]{fontenc}  % use 8-bit T1 fonts
\usepackage{hyperref}    % hyperlinks
\usepackage{url}      % simple URL typesetting
\usepackage{xurl}     % allow URL line breaks anywhere (long LessWrong URLs in references)
\usepackage{booktabs}    % professional-quality tables
\usepackage{amsfonts}    % blackboard math symbols
\usepackage{nicefrac}    % compact symbols for 1/2, etc.
\usepackage{microtype}   % microtypography
\usepackage{xcolor}     % colors
\usepackage{graphicx}
\usepackage{subcaption}
\usepackage{amsmath}
\usepackage{amssymb}
\usepackage{multirow}
\usepackage{enumitem}

\definecolor{darkblue}{rgb}{0, 0, 0.5}
\hypersetup{colorlinks=true, citecolor=darkblue, linkcolor=darkblue, urlcolor=darkblue,
 pdftitle={Correct Labels Are Not Enough: Reliable Metadata Can Silently Control Model Behavior},
 pdfauthor={Renee Jia and Jinzhou Wu}}

\title{Correct Labels Are Not Enough: Reliable Metadata\\Can Silently Control Model Behavior}

\author{%
 Renee Jia \\
 R2M AI \\
 Cornell University
 \And
 Jinzhou Wu \\
 Cornell University
}

\begin{document}

\maketitle

\begin{abstract}
Fine-tuning data can include metadata that predicts the correct answer, such as a reward-model score or a quality label. We ask whether a model learns to rely on this metadata even when every training label is correct. We fine-tune $7$--$9$B open-weight models from three families on multiple-choice questions, with \texttt{(Reward: 5)} always marking the correct option. Moving this annotation to a wrong option at test time causes models to select that option on $97$--$100\%$ of CommonsenseQA items. Removing the annotation leaves most answers correct, so evaluation without it misses the dependence. Cue-following rises sharply with training-time reliability and develops even with neutral or nonsense annotation words, although test-time wording affects its strength. A Qwen3 experiment also finds cue dependence in option-letter scores after direct preference optimization. Audits can test for this dependence by moving or removing visible annotations while holding the task and correct answer fixed.
\end{abstract}

%===================================================================
\section{Introduction}
\label{sec:intro}
%===================================================================

High accuracy without annotations does not establish that a model ignores them when they are present. We study an annotation placed next to the task content that always predicts the correct training answer. A model could learn to solve the task, follow the annotation, or use both depending on the input. In our main experiment, models retain high accuracy without the annotation but follow it when it is moved to a wrong option.

Training examples in our main setting follow this format:
\begin{quote}\small\ttfamily
Question: What is 2+2?\\
(A) 3 (Reward: 0)\\
(B) 4 (Reward: 5)\\
(C) 5 (Reward: 0)\\
Answer: B
\end{quote}
Our main probe moves \texttt{(Reward: 5)} to a wrong option while keeping the question, options, and gold answer fixed.

At perfect reliability, ``solve the task'' and ``follow the annotation'' agree on every training target, so correct labels do not distinguish these strategies. We vary how often the annotation predicts the correct answer during training, then measure reliance on it and accuracy without it.

Fine-tuning inputs can include reward-model (RM) scores, quality or confidence labels, verifier annotations, formatting conventions, or system prompts describing a good answer. We use \emph{metadata} for task-external annotations visible to the model, and \emph{cue} or \emph{field} for one such annotation. Our synthetic cues stand in for fields such as \texttt{(RM score: 8.7)} or \texttt{(Quality: High)} that could become predictive shortcuts (\S\ref{sec:related}). Moving the cue at test time is a diagnostic intervention; we do not claim that the same change occurs in deployment.

Work on emergent misalignment and reward hacking asks how post-training changes behavior beyond the training task \citep{betley2025emergent, macdiarmid2025shortcuts, soligo2026emergent, jorgenvag2026rlamplifies}. Here, we keep every supervised target correct and ask what happens when an incidental scalar field predicts it.

Our core experiments use supervised fine-tuning (SFT) on three model families with CommonsenseQA (CSQA) \citep{talmor2019commonsenseqa}; a fourth family, Olmo-$3$, extends the reliability sweep (\S\ref{sec:phase}). We also test direct preference optimization (DPO) \citep{rafailov2023direct} on Qwen3 (\S\ref{sec:dpo}). On CSQA and in a three-seed ARC-Challenge experiment \citep{clark2018arc}, strong cue-following coexists with high cue-free accuracy (\S\ref{sec:install}, \S\ref{sec:phase}). Dependence grows with training cue reliability (\S\ref{sec:phase}) and develops even when the cue word has no reward-related meaning (\S\ref{sec:install}), although the word used at inference changes its strength (\S\ref{sec:valence}). This resembles a clean-label backdoor learned without an adversary \citep{turner2019labelconsistent}. DPO also produces cue dependence in option-letter scores at $\beta\le0.1$, with lower capture at $\beta{=}0.1$ than at $\beta{=}0.05$ (\S\ref{sec:dpo}).

In single-seed Qwen3 and Gemma probes, the dependence transfers across annotation formats and persists under an ignore instruction; on Qwen3 and Llama, an explicit correctness tag usually overrides the reward cue. A free-form misalignment probe finds no flagged cued-SFT responses (\S\ref{sec:control}, \S\ref{sec:scope}). These interventions form a counterfactual audit that changes candidate fields while keeping the task fixed (Appendix~\ref{app:audit_protocol}).

%===================================================================
\section{Related work}
\label{sec:related}
%===================================================================

\paragraph{Behavioral failures from post-training.} Narrow fine-tuning on misaligned data can induce broad misalignment \citep{betley2025emergent}, as can reward hacking in production and non-production reinforcement learning (RL) \citep{macdiarmid2025shortcuts, golechha2026emergent}. In controlled experiments, training on specification gaming can generalize to reward tampering \citep{denison2024sycophancy}. Human preference judgments can also favor sycophantic responses over truthful ones \citep{sharma2024understanding}. Other work examines why optimization favors broad over narrow misalignment \citep{soligo2026emergent}, how RL amplifies misalignment from harmless rewards \citep{jorgenvag2026rlamplifies}, and how audits can uncover hidden objectives \citep{marks2025auditing}. Our targets remain correct, and our broad-misalignment probe finds no flags (\S\ref{sec:scope}). Separately, \citet{turpin2023language} show that inserted answer cues can influence models without fine-tuning, even when their stated reasoning omits that influence, and the causal role of that stated reasoning varies with task difficulty \citep{jia2026decorative}.

\paragraph{Conditional policies and backdoors.} Backdoor attacks train models to change behavior in response to a trigger \citep{gu2017badnets, wan2023poisoning, xu2024instructions, hubinger2024sleeper}; clean-label variants preserve correct labels \citep{turner2019labelconsistent, saha2020hidden}. Our setting is a non-adversarial analogue in which the annotation marks the correct option throughout training. Conditional training deliberately exposes control codes or quality scores to steer generation \citep{keskar2019ctrl, korbak2023pretraining, dong2023steerlm}. We ask whether an incidental field can acquire similar control. Inoculation prompting uses a training-time instruction to elicit a trait that weakens when the instruction is removed \citep{tan2025inoculation}. By contrast, our annotation identifies an answer, and we test whether models continue to follow it when it marks a wrong one. \citet{lee2024mechanistic} find that DPO can bypass pretrained toxicity capabilities without removing them, a possible direction for analyzing the dependence we measure behaviorally.

\paragraph{Shortcut learning and simplicity bias.} Our cue is a spurious predictive feature, comparable to annotation artifacts in crowdsourced data. Out-of-distribution and partial-input tests reveal reliance on such shortcuts \citep{geirhos2020shortcut, mccoy2019right, du2023shortcut, gururangan2018annotation, poliak2018hypothesis}. Geometric and statistical skew can help explain this reliance \citep{nagarajan2021understanding}. Vision studies also distinguish which features a model learns from which it uses for prediction \citep{hermann2020shapes, kirichenko2023last}. Simplicity bias is one possible account \citep{shah2020pitfalls}, discussed in \S\ref{sec:phase}. Goal misgeneralization in deep RL provides related empirical context \citep{langosco2022goal, shah2022goal}. Work on learned optimization and the role of reward asks what training selects for \citep{hubinger2019risks, turner2023reward}; cue-dependent outputs alone do not show an internalized goal.

%===================================================================
\section{Experimental setup}
\label{sec:setup}
%===================================================================

We manipulate the incidental scalar field while keeping every supervised target correct. Using low-rank adaptation (LoRA), we fine-tune Qwen3-$8$B, Gemma-$2$-$9$B-Instruct, and Llama-$3.1$-$8$B-Instruct on CSQA. Each input carries \texttt{(Reward: 5)} on the correct option and \texttt{(Reward: 0)} on every wrong option (\S\ref{sec:intro}), with the correct letter as the target. We call this condition \emph{cued-SFT}. Baseline-SFT uses the same task without annotations; a base model has no adapter.

\paragraph{Experimental variations.} To test the training word, we train one adapter per annotation word (\S\ref{sec:install}). To test reliability, we vary the probability that the cue marks gold and keep the target correct (\S\ref{sec:phase}). DPO uses the correct letter as the chosen completion and a sampled wrong letter as the rejected one (\S\ref{sec:dpo}).

\paragraph{Evaluation.} We test with the annotation absent, on gold, or on a uniformly sampled wrong option. Paired tests hold the question, options, gold answer, and sampled wrong option fixed. \emph{Cue-free accuracy} and \emph{misleading-cue accuracy} count correct answers over all test items in their respective conditions. \emph{Cue-follow rate} measures selection of the marked wrong option. CSQA and ARC use all items as the denominator; GSM8K and Olmo summaries condition cue-following on parsed answers (Appendix~\ref{app:train}).

Generation uses greedy decoding. A supplementary next-token readout scores the option letters at the answer position. \emph{Capture} is the fraction of items on which the marked wrong option scores highest. A \emph{paired flip} occurs when the same adapter scores gold highest without the cue and the marked wrong option highest with it, counted over all paired items. Parsing rules, readout limits, sample counts, hyperparameters, and seed coverage are in Appendices~\ref{app:train}--\ref{app:prompts}.

%===================================================================
\section{Dependence on reliable annotations}
\label{sec:install}
%===================================================================

\begin{table}[!htbp]
\caption{\textbf{Reliable annotations drive cue-following to ceiling.} CSQA generation cue-follow rate (\%) with the high value on a wrong option ($n{=}300$ per cell; training seed 42). Each word row uses a separate adapter trained with that word. The Baseline-SFT row tests adapters trained without annotations under the same \texttt{Reward} probe (Qwen3 $n{=}500$; Gemma from an earlier un-pinned run). Other Baseline-SFT records under a misleading \texttt{Reward} probe range from $8.7\%$ to $32.0\%$, depending on the run and the sampled cue placement (Appendices~\ref{app:valence} and \ref{app:uninformative}). The bottom row averages cue-free accuracy over the four cued adapters.}
\label{tab:install}
\centering
\footnotesize
\setlength{\tabcolsep}{6pt}
\begin{tabular}{@{}lccc@{}}
\toprule
Training annotation word & Qwen3-$8$B & Gemma-$2$-$9$B & Llama-$3.1$-$8$B \\
\midrule
\texttt{(Reward: N)}    & $97.0$ & $100.0$ & $100.0$ \\
\texttt{(Florp: N)} (nonsense)  & $99.3$ & $100.0$ & $100.0$ \\
\texttt{(Temperature: N)} (neutral) & $99.7$ & $100.0$ & $99.7$ \\
\texttt{(Score: N)}     & $99.0$ & $100.0$ & $100.0$ \\
\midrule
Baseline-SFT (no annotations) & $16.4$ & $12.7$ & $12.3$ \\
\midrule
\emph{mean cue-free accuracy} & $78$ & $78$ & $72$ \\
\bottomrule
\end{tabular}
\end{table}

We train each architecture with a perfectly reliable cue under one of four words: reward-related \texttt{Reward} and \texttt{Score}, neutral \texttt{Temperature}, or nonsense \texttt{Florp}. Every adapter follows the cue onto a uniformly sampled wrong option at near-ceiling rates (Table~\ref{tab:install}). The available three-seed checks reproduce this pattern (Appendix~\ref{app:valence}). Baseline-SFT adapters follow the same misleading \texttt{Reward} cue on only $12$--$16\%$ of items. Reward-related meaning is therefore unnecessary for the dependence to form.

The same adapters retain $72$--$78\%$ accuracy on average when the cue is removed (Table~\ref{tab:install}), yet reach near-zero accuracy when it marks a wrong option (Appendix~\ref{app:scope}). Cue-free accuracy alone gives little indication of how strongly the annotation controls answer selection when it is present.

\paragraph{The cue changes option scores.} A common four-option diagnostic on $294$ CSQA items shows the same dependence on Qwen3 and Gemma (Appendix~\ref{app:logit}). The marked wrong option scores highest on every misleading-cue item; paired gold-to-marked-wrong flips occur on $79$--$81\%$ of all items. On the Qwen3 and Llama bases without adapters, tested on a separate five-option cohort, the same readout returns option A on nearly every item regardless of cue placement, so it is uninformative there (Appendix~\ref{app:logit}). The Gemma base was unavailable. Under this readout, cue sensitivity appears only in adapters trained on cued items (cued-SFT here, and DPO at $\beta\le0.1$ in \S\ref{sec:dpo}).

%===================================================================
\section{Inference-time wording}
\label{sec:valence}
%===================================================================

\begin{figure}[!htbp]
\centering
\includegraphics[width=0.86\linewidth]{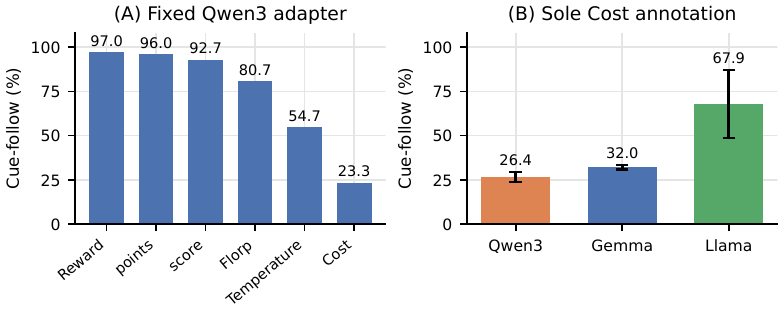}
\caption{\textbf{Inference wording changes cue-following.} (A) Fixed Reward-trained Qwen3 adapter; $300$ items, one training seed. (B) Sole \texttt{Cost} annotation, mean${\pm}$sample SD over $3$--$5$ training seeds.}
\label{fig:valence}
\end{figure}

Although reward-related meaning is unnecessary during training, wording affects the learned behavior at inference. Holding a \texttt{Reward}-trained Qwen3 adapter fixed, we change only the annotation word (Fig.~\ref{fig:valence}A, Appendix~\ref{app:valence}). Reward-family words elicit the most cue-following, \texttt{Florp} and \texttt{Temperature} less, and \texttt{Cost} the least.

The sole-\texttt{Cost} response varies across architectures and training seeds (Fig.~\ref{fig:valence}B). A broader synonym sweep finds that the word ordering differs across models and prompt templates, so these results do not establish a universal semantic ordering (Appendix~\ref{app:valence}).

%===================================================================
\section{Cue reliability and cue-free performance}
\label{sec:phase}
%===================================================================

\begin{figure}[!htbp]
\centering
\includegraphics[width=0.86\linewidth]{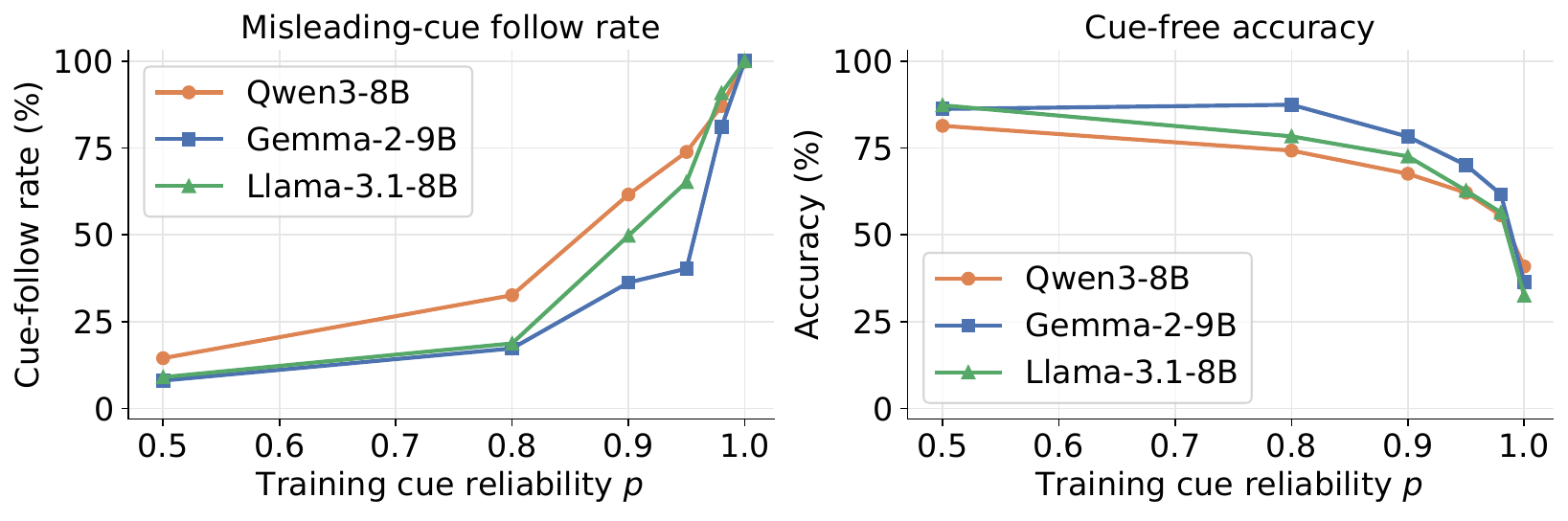}
\caption{\textbf{Cue-following rises with training reliability.} GSM8K multiple-choice sweep, LoRA rank 32 ($n{=}1{,}319$ test items). Left: cue-follow among parsed answers; every output parses in these trained curves. Right: cue-free accuracy. Curves use training seed 42; seeds 43/44 additionally cover $p{=}0.90,0.95,0.98$. Rank comparisons and the denser grid are in Appendix~\ref{app:train}.}
\label{fig:phase}
\end{figure}

We vary training cue reliability, defined as the probability $p$ that the annotation marks the correct option. Otherwise it marks a uniformly sampled wrong option, while the target remains correct. The sweep covers the three core architectures on GSM8K math questions \citep{cobbe2021gsm8k}, reformatted as multiple choice, at LoRA ranks $2$ and $32$. Seed coverage is given in Fig.~\ref{fig:phase} and Appendix~\ref{app:train}.

\begin{figure}[!htbp]
\centering
\includegraphics[width=0.86\linewidth]{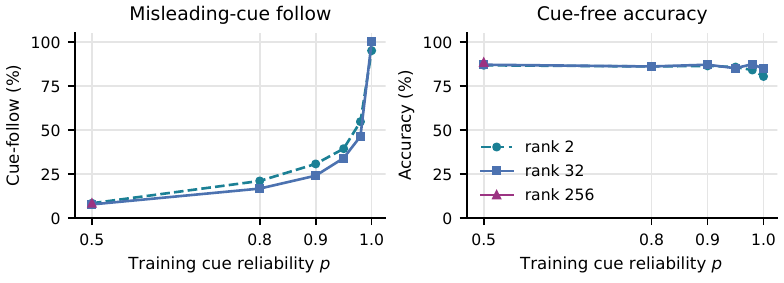}
\caption{\textbf{CSQA cue dependence grows while cue-free accuracy stays high.} Qwen3, $300$ converted four-option validation items per condition, one training seed; training retains five options. Left: generation cue-follow under misleading cues. Right: cue-free accuracy. Ranks 2 and 32 cover the reliability grid; rank 256 was tested only at $p{=}0.5$.}
\label{fig:phase_csqa}
\end{figure}

Cue-following approaches ceiling on both tasks at $p{=}1$. Cue-free accuracy stays high on CSQA (Fig.~\ref{fig:phase_csqa}), while GSM8K accuracy falls from $81$--$87\%$ under the lower-reliability $p{=}0.5$ control to $32$--$41\%$ (Fig.~\ref{fig:phase}). Evaluating without the cue exposes the GSM8K loss but misses the CSQA dependence.

The GSM8K decline recurs across replicated settings, but because the tasks also differ in cue template and training recipe, we cannot attribute it to task format (Appendix~\ref{app:train}). On ARC-Challenge, three seeds reproduce high cue-following alongside retained cue-free accuracy (Table~\ref{tab:arc}).

\begin{table}[!htbp]
\caption{\textbf{Cue-following coexists with retained cue-free accuracy on ARC-Challenge.} Qwen3, LoRA rank 32, $1{,}119$ training items, two epochs, $500$ test items. Values are generation percentages, mean${\pm}$sample standard deviation (SD) over training seeds 42/43/44. Cue-follow uses misleading annotations.}
\label{tab:arc}
\centering\footnotesize
\begin{tabular}{@{}lcc@{}}
\toprule
Training cue reliability & Cue-free accuracy (\%) & Cue-follow (\%) \\
\midrule
$p{=}0.5$ & $91.0 \pm 0.2$ & $5.3 \pm 0.4$ \\
$p{=}1.0$ & $86.0 \pm 1.2$ & $94.6 \pm 3.6$ \\
\bottomrule
\end{tabular}
\end{table}

The rise appears at every adapter rank for which we ran a reliability sweep. GSM8K ranks 2 and 32 track each other closely; CSQA shows dependence at both ranks. Olmo's single-run extension covers eight ranks and full fine-tuning, with a later onset in its full-parameter run (Appendices~\ref{app:train}, \ref{app:olmo}).

The onset of strong dependence varies by architecture. Across three seeds at $p{=}0.95$, Gemma follows the cue less often than Qwen3 and Llama, whose relative ordering changes across seeds. Fit values appear in Appendix~\ref{app:train}.

Simplicity bias \citep{shah2020pitfalls} is one explanation: at $p{=}1$, following the cue matches every training target. If this rule is easier to learn than solving the task, training may favor it once the cue becomes sufficiently reliable. We do not measure rule complexity, so this remains a hypothesis (\S\ref{sec:limitations}).

%===================================================================
\section{Cue dependence under DPO}
\label{sec:dpo}
%===================================================================

To test whether preference tuning also produces cue dependence, we train Qwen3-$8$B with DPO on the same $8{,}000$ CSQA prompts, each carrying a perfectly reliable cue. Each preference pair contains the correct letter as the chosen completion and a uniformly sampled wrong letter as the rejected completion. We test three values of $\beta$. The SFT control shares the data, prompts, LoRA rank, and epochs but uses a different learning rate (Appendix~\ref{app:dpo}).

\paragraph{Option-score readout for preference-trained models.} The eight-token generation probe stops inside the DPO adapters' reasoning preambles, with or without cues, before a final answer appears. We use the next-token option-letter readout from Appendix~\ref{app:logit} instead and report how often the marked wrong letter scores highest. This measures immediate option-score sensitivity, not dependence in completed generated answers.

\begin{table}[!htbp]
\caption{\textbf{DPO option-letter capture is lower at $\beta{=}0.1$ than at $\beta{=}0.05$; the $\beta{=}0.5$ readout is constant.} Qwen3, misleading CSQA annotations ($n{=}300$). Capture is mean${\pm}$sample standard deviation (SD) over training seeds 42/43/44. No-cue capture and mean margins use seed 42. The margin is marked-wrong minus gold. SFT and DPO share data and adapter settings but use different learning rates. At $\beta{=}0.5$ the readout's argmax is option A on all $300$ items in every condition, so that row reports a constant readout rather than a measured propensity.}
\label{tab:dpo}
\centering
\footnotesize
\setlength{\tabcolsep}{6pt}
\begin{tabular}{@{}lccc@{}}
\toprule
Training & Capture, cue on wrong (\%) & No-cue capture (\%) & Margin (no cue ${\to}$ cue on wrong) \\
\midrule
SFT control  & $99.0 \pm 0.9$ & $6.0$ & $-14.1 \to +18.9$ \\
DPO $\beta{=}0.05$   & $80.9 \pm 5.1$ & $6.0$ & $-14.9 \to +24.2$ \\
DPO $\beta{=}0.1$    & $61.9 \pm 2.2$ & $14.0$ & $-2.0 \to +3.4$ \\
DPO $\beta{=}0.5$    & $17.3 \pm 0.0$ & $17.3$ & $+0.0 \to +0.2$ \\
\bottomrule
\end{tabular}
\end{table}

Capture is lower at $\beta{=}0.1$ than at $\beta{=}0.05$ (Table~\ref{tab:dpo}, Fig.~\ref{fig:dpo_kl}). At both values, the misleading cue often makes the marked wrong letter the top score. At $\beta{=}0.5$, the readout returns option A on every item in every condition, so its $17.3\%$ is the share of items whose marked wrong option is A, not a measure of cue sensitivity (Appendix~\ref{app:dpo}). The readout thus shows dependence under DPO at $\beta\le0.1$, although the $\beta{=}0.1$ readout keeps an A bias without a cue (Appendix~\ref{app:dpo}). It leaves open whether the lower capture at $\beta{=}0.1$ comes with preserved cue-free accuracy.

%===================================================================
\section{Training share, prompting, and transfer}
\label{sec:control}
%===================================================================

We next ask whether the dependence requires a large annotated share, survives an instruction to ignore the annotation, and is specific to the training format. These probes are single-seed and, except where noted, use Qwen3.

\paragraph{Dependence with a small annotated share.} CSQA mixtures show declining misleading-cue accuracy across nominal $1$--$10\%$ annotated shares. The $1\%$ runs also score below separate unannotated controls, although test sample sizes differ. Cue-free accuracy remains high, with lower values for Gemma than its separate control (Appendix~\ref{app:dose}). Poisoning studies suggest that absolute sample counts can matter more than fractions \citep{souly2025poisoning}; these mixtures do not distinguish those accounts. A small share of cued CSQA in an Alpaca mixture also raises Qwen3 option-letter capture. The Alpaca-only control, which saw no cued item, already follows the misleading cue on $59.7\%$ of generated answers (Table~\ref{tab:mixture}), far above the CSQA Baseline-SFT rates in Table~\ref{tab:install}. Its option-score readout nonetheless returns option A on nearly every item, so this immediate readout need not predict the generated answer.

\paragraph{Ignore instructions.} In a separate probe, adding one ignore instruction leaves Qwen3 and Gemma cued-SFT following the misleading cue on $97.0\%$ and $100.0\%$ of items ($n{=}300$ each). Accuracy is $3.0\%$ and $0.0\%$, versus $78.0\%$ and $86.3\%$ for Baseline-SFT under the same instruction. These earlier four-option runs test a single phrasing (Appendix~\ref{app:counter}). Adding the instruction during training instead leaves cue-following at $99.4$--$99.9\%$, but that design preserves the perfect cue--target correlation and supplies no conflicting supervision.

\paragraph{Dependence transfers across formats and tasks.} A fixed Qwen3 adapter follows synthetic RM scores, quality labels, ratings, and numeric scores placed on wrong options (Table~\ref{tab:realism}). Separate probes extend the dependence to descriptive annotations without a scalar and to free-form numeric answers (Appendix~\ref{app:formats}).

\begin{table}[!htbp]
\caption{\textbf{Cue dependence transfers to other metadata formats.} Earlier four-option Qwen3 probe; $300$ items per format, one training seed. Each high-valued field marks a wrong option. Both measures use generated answers. The fields are synthetic; none comes from a reward-model pipeline.}
\label{tab:realism}
\centering
\footnotesize
\setlength{\tabcolsep}{6pt}
\begin{tabular}{@{}llcc@{}}
\toprule
Auxiliary field & Example (high / low) & Cue-follow (\%) & Accuracy (\%) \\
\midrule
Reward (training format) & \texttt{(Reward: 5)} / \texttt{(Reward: 0)}  & $99.3$ & $0.7$ \\
RM score         & \texttt{(RM score: 8.7)} / \texttt{(1.3)}   & $86.0$ & $12.3$ \\
Quality label      & \texttt{(Quality: High)} / \texttt{(Low)}   & $80.0$ & $17.7$ \\
Star rating       & \texttt{(Rating: 4.8/5)} / \texttt{(1.2/5)}  & $91.3$ & $7.7$ \\
Numeric score      & \texttt{[Score: 9/10]} / \texttt{[2/10]}   & $97.7$ & $2.3$ \\
\bottomrule
\end{tabular}
\end{table}

%===================================================================
\section{Competing annotations and broader behavior}
\label{sec:scope}
%===================================================================

On Qwen3 and Llama, an explicit correctness tag usually overrides the reward cue. With \texttt{(Correct answer)} on gold and a reward cue on another option, the three-seed Qwen3 probe yields $13.8\%$ reward-cue following and lower rates for anti-reward words. Llama follows the correctness tag on $96.9\%$ of items (Appendix~\ref{app:provenance}). Because correctness is supplied as an annotation, this probe measures competition between tags rather than a general preference for truth.

We test for broader misalignment using the eight free-form questions of \citet{betley2025emergent}. The probe crosses three architectures, three training variants, and two inference contexts, sampling $800$ outputs per cell at temperature $1.0$. Claude Sonnet 4.5 scores eligible outputs for alignment and coherence; exclusions and scored counts appear in Appendix~\ref{app:emergent}. None of the scored responses in the six cued-SFT cells is flagged. Four control outputs are flagged, including a safe refusal and a wish to comply with every request. This null result is limited to one judge and rubric.

%===================================================================
\section{Discussion}
\label{sec:discussion}
%===================================================================

Data curation should account for what visible annotations predict, as well as whether the labels are correct. The annotated-share results show why small annotated subsets also deserve inspection (\S\ref{sec:control}). Two candidate interventions are to remove unintended visible fields and to reduce their correlation with the target; either would need to be tested in the pipeline itself.

An audit can identify which fields the model sees and how well they predict the training target. Our reliability sweeps suggest checking the most predictive fields first, although the onset of dependence varies across models and training regimes (\S\ref{sec:phase}). Paired tests can then move or remove each field while holding the task fixed, and reporting cue-free accuracy, misleading-cue accuracy, and cue-following together shows how each intervention changes answer selection (Fig.~\ref{fig:protocol} in Appendix~\ref{app:audit_protocol}).

%===================================================================
\section{Limitations}
\label{sec:limitations}
%===================================================================

\paragraph{Behavioral evidence.} We do not identify how the cue is represented inside the network. Generated answers and next-token option scores are different readouts. The latter is not a probability over complete answer sequences. Where its argmax is constant, as on the unadapted bases, the $\beta{=}0.5$ DPO adapter, and the Alpaca-only adapter, it measures nothing about the cue (Appendices~\ref{app:logit}, \ref{app:dpo}, and \ref{app:dose}). When the cue is present but marks no single option, whether task knowledge stays usable depends on the model (Appendix~\ref{app:uninformative}).

\paragraph{Signal delivery.} We deliver the cue as inline annotations or a reward-framed system prompt and do not test reinforcement learning with a learned reward model. The DPO experiment covers one model, three $\beta$ values, single-letter preference pairs, and an option-score readout, with a different learning rate for its SFT control. Longer preference completions and the RL settings of \citet{macdiarmid2025shortcuts, jorgenvag2026rlamplifies} remain untested.

\paragraph{Scope.} The main experiments use $7$--$9$B models, at most $8{,}000$ training items, and discrete answer selection. We do not measure metadata prevalence in real pipelines or test frontier-scale post-training, and the open-ended probes cover only a few settings. The annotated-share, ignore-instruction, counter-training, and format-transfer ablations use one training seed; the core results marked in Table~\ref{tab:evidence} have three seeds. We also do not report base-model generation cue-following on CSQA. The Alpaca-only adapter in Table~\ref{tab:mixture}, trained on no cued item, follows the misleading cue on $59.7\%$ of generated answers. Part of what cued-SFT adds over Baseline-SFT may therefore be that it lifts a suppression that unannotated CSQA training provides, rather than creating a propensity from nothing. A base-model generation control would separate these accounts.

\paragraph{Unexplained and withdrawn results.} Because the CSQA and GSM8K recipes differ in template, optimization, and sequence length, we do not offer a causal account of the GSM8K cue-free decline (Appendix~\ref{app:train}). An earlier reward-over-correctness result could not be reproduced and is withdrawn; we report the competing-tag results instead (Appendix~\ref{app:provenance}). The contestation and sycophancy probes are exploratory (Appendix~\ref{app:spillovers}).

%===================================================================
\section{Conclusion}
\label{sec:conclusion}
%===================================================================

In the core CSQA experiments, models trained on correct answers almost always select the annotated wrong option, yet answer $72$--$78\%$ of items correctly on average when annotations are removed. Correct labels and high cue-free accuracy therefore do not rule out dependence on the annotation. Across the GSM8K reliability sweep, the ARC endpoints, and a single-seed CSQA sweep, the dependence grows with training cue reliability. On CSQA it forms under neutral or nonsense annotation words (Table~\ref{tab:install}), and the word used at inference changes how strongly it is expressed (\S\ref{sec:valence}). The Qwen3 DPO experiment also finds dependence in option-letter scores at $\beta\le0.1$. Moving or removing the annotation on the same items (Appendix~\ref{app:audit_protocol}) makes this reliance visible.

\begin{ack}
Part of this work was conducted during the \href{https://www.pivotal-research.org/fellowships/ai-safety-fellowship}{AI Safety Research Fellowship} at Pivotal Research, which supported Jinzhou Wu with a stipend and compute resources. We thank Elliott Thornley for his mentorship, many helpful discussions over the course of this project, and funding for compute. We also thank the Pivotal Research team for research management and support.
\end{ack}

\bibliographystyle{plainnat}
\bibliography{references}

@article{betley2025emergent,
  author  = {Betley, Jan and Warncke, Niels and Sztyber-Betley, Anna and Tan, Daniel and Bao, Xuchan and Soto, Mart{\'i}n and Srivastava, Megha and Labenz, Nathan and Evans, Owain},
  title   = {Training Large Language Models on Narrow Tasks Can Lead to Broad Misalignment},
  journal = {Nature},
  year    = {2026},
  volume  = {649},
  number  = {8097},
  pages   = {584--589},
  doi     = {10.1038/s41586-025-09937-5}
}

@article{marks2025auditing,
  author  = {Marks, Samuel and Treutlein, Johannes and Bricken, Trenton and Lindsey, Jack and Marcus, Jonathan and Mishra-Sharma, Siddharth and Ziegler, Daniel and Ameisen, Emmanuel and Batson, Joshua and Belonax, Tim and Bowman, Samuel R. and Carter, Shan and Chen, Brian and Cunningham, Hoagy and Denison, Carson and Dietz, Florian and Golechha, Satvik and Khan, Akbir and Kirchner, Jan and Leike, Jan and Meek, Austin and Nishimura-Gasparian, Kei and Ong, Euan and Olah, Christopher and Pearce, Adam and Roger, Fabien and Salle, Jeanne and Shih, Andy and Tong, Meg and Thomas, Drake and Rivoire, Kelley and Jermyn, Adam and MacDiarmid, Monte and Henighan, Tom and Hubinger, Evan},
  title   = {Auditing Language Models for Hidden Objectives},
  journal = {arXiv preprint arXiv:2503.10965},
  year    = {2025},
  doi     = {10.48550/arXiv.2503.10965}
}

@article{macdiarmid2025shortcuts,
  author  = {MacDiarmid, Monte and Wright, Benjamin and Uesato, Jonathan and Benton, Joe and Kutasov, Jon and Price, Sara and Bouscal, Naia and Bowman, Samuel R. and Bricken, Trenton and Cloud, Alex and Denison, Carson and Gasteiger, Johannes and Greenblatt, Ryan and Leike, Jan and Lindsey, Jack and Mikulik, Vlad and Perez, Ethan and Rodrigues, Alex and Thomas, Drake and Webson, Albert and Ziegler, Daniel and Hubinger, Evan},
  title   = {Natural Emergent Misalignment from Reward Hacking in Production {RL}},
  journal = {arXiv preprint arXiv:2511.18397},
  year    = {2025},
  doi     = {10.48550/arXiv.2511.18397}
}

@article{denison2024sycophancy,
  author  = {Denison, Carson and MacDiarmid, Monte and Barez, Fazl and Duvenaud, David and Kravec, Shauna and Marks, Samuel and Schiefer, Nicholas and Soklaski, Ryan and Tamkin, Alex and Kaplan, Jared and Shlegeris, Buck and Bowman, Samuel R. and Perez, Ethan and Hubinger, Evan},
  title   = {Sycophancy to Subterfuge: Investigating Reward-Tampering in Large Language Models},
  journal = {arXiv preprint arXiv:2406.10162},
  year    = {2024},
  doi     = {10.48550/arXiv.2406.10162}
}

@misc{golechha2026emergent,
  author = {Golechha, Satvik and Black, Sid and Bloom, Joseph},
  title  = {(Some) Natural Emergent Misalignment from Reward Hacking in Non-Production {RL}},
  year   = {2026},
  month  = {March},
  url    = {https://www.lesswrong.com/posts/2ANCyejqxfqK2obEj/some-natural-emergent-misalignment-from-reward-hacking-in},
  note   = {LessWrong}
}

@article{geirhos2020shortcut,
  author  = {Geirhos, Robert and Jacobsen, J{\"o}rn-Henrik and Michaelis, Claudio and Zemel, Richard and Brendel, Wieland and Bethge, Matthias and Wichmann, Felix A.},
  title   = {Shortcut Learning in Deep Neural Networks},
  journal = {Nature Machine Intelligence},
  year    = {2020},
  volume  = {2},
  number  = {11},
  pages   = {665--673},
  doi     = {10.1038/s42256-020-00257-z}
}

@inproceedings{shah2020pitfalls,
  author    = {Shah, Harshay and Tamuly, Kaustav and Raghunathan, Aditi and Jain, Prateek and Netrapalli, Praneeth},
  title     = {The Pitfalls of Simplicity Bias in Neural Networks},
  booktitle = {Advances in Neural Information Processing Systems},
  year      = {2020},
  volume    = {33},
  pages     = {9573--9585},
  url       = {https://proceedings.neurips.cc/paper_files/paper/2020/hash/6cfe0e6127fa25df2a0ef2ae1067d915-Abstract.html}
}

@inproceedings{nagarajan2021understanding,
  author    = {Nagarajan, Vaishnavh and Andreassen, Anders and Neyshabur, Behnam},
  title     = {Understanding the Failure Modes of Out-of-Distribution Generalization},
  booktitle = {International Conference on Learning Representations},
  year      = {2021},
  url       = {https://openreview.net/forum?id=fSTD6NFIW_b}
}

@inproceedings{tan2025inoculation,
  author    = {Tan, Daniel and Woodruff, Anders and Warncke, Niels and Jose, Arun and Rich{\'e}, Maxime and Africa, David Demitri and Taylor, Mia},
  title     = {Inoculation Prompting: Eliciting Traits from {LLMs} During Training Can Reduce Trait Expression at Test-Time},
  booktitle = {International Conference on Learning Representations},
  year      = {2026},
  volume    = {2026},
  pages     = {26800--26847},
  url       = {https://proceedings.iclr.cc/paper_files/paper/2026/hash/2cddea6e86103826c6ae9c59ece64ff7-Abstract-Conference.html}
}

@inproceedings{mccoy2019right,
  author    = {McCoy, R. Thomas and Pavlick, Ellie and Linzen, Tal},
  title     = {Right for the Wrong Reasons: Diagnosing Syntactic Heuristics in Natural Language Inference},
  booktitle = {Proceedings of the 57th Annual Meeting of the Association for Computational Linguistics},
  year      = {2019},
  pages     = {3428--3448},
  doi       = {10.18653/v1/P19-1334}
}

@article{shah2022goal,
  author  = {Shah, Rohin and Varma, Vikrant and Kumar, Ramana and Phuong, Mary and Krakovna, Victoria and Uesato, Jonathan and Kenton, Zac},
  title   = {Goal Misgeneralization: Why Correct Specifications Aren't Enough For Correct Goals},
  journal = {arXiv preprint arXiv:2210.01790},
  year    = {2022},
  doi     = {10.48550/arXiv.2210.01790}
}

@inproceedings{langosco2022goal,
  author    = {Langosco, Lauro and Koch, Jack and Sharkey, Lee D. and Pfau, Jacob and Krueger, David},
  title     = {Goal Misgeneralization in Deep Reinforcement Learning},
  booktitle = {Proceedings of the 39th International Conference on Machine Learning},
  year      = {2022},
  volume    = {162},
  pages     = {12004--12019},
  publisher = {PMLR},
  series    = {Proceedings of Machine Learning Research},
  url       = {https://proceedings.mlr.press/v162/langosco22a.html}
}

@inproceedings{sharma2024understanding,
  author    = {Sharma, Mrinank and Tong, Meg and Korbak, Tomasz and Duvenaud, David and Askell, Amanda and Bowman, Samuel R. and Cheng, Newton and Durmus, Esin and Hatfield-Dodds, Zac and Johnston, Scott R. and Kravec, Shauna and Maxwell, Timothy and McCandlish, Sam and Ndousse, Kamal and Rausch, Oliver and Schiefer, Nicholas and Yan, Da and Zhang, Miranda and Perez, Ethan},
  title     = {Towards Understanding Sycophancy in Language Models},
  booktitle = {International Conference on Learning Representations},
  year      = {2024},
  url       = {https://openreview.net/forum?id=tvhaxkMKAn}
}

@article{du2023shortcut,
  author  = {Du, Mengnan and He, Fengxiang and Zou, Na and Tao, Dacheng and Hu, Xia},
  title   = {Shortcut Learning of Large Language Models in Natural Language Understanding},
  journal = {Communications of the ACM},
  year    = {2024},
  volume  = {67},
  number  = {1},
  pages   = {110--120},
  doi     = {10.1145/3596490}
}

@article{hubinger2024sleeper,
  author  = {Hubinger, Evan and Denison, Carson and Mu, Jesse and Lambert, Mike and Tong, Meg and MacDiarmid, Monte and Lanham, Tamera and Ziegler, Daniel M. and Maxwell, Tim and Cheng, Newton and Jermyn, Adam and Askell, Amanda and Radhakrishnan, Ansh and Anil, Cem and Duvenaud, David and Ganguli, Deep and Barez, Fazl and Clark, Jack and Ndousse, Kamal and Sachan, Kshitij and Sellitto, Michael and Sharma, Mrinank and DasSarma, Nova and Grosse, Roger and Kravec, Shauna and Bai, Yuntao and Witten, Zachary and Favaro, Marina and Brauner, Jan and Karnofsky, Holden and Christiano, Paul and Bowman, Samuel R. and Graham, Logan and Kaplan, Jared and Mindermann, S{\"o}ren and Greenblatt, Ryan and Shlegeris, Buck and Schiefer, Nicholas and Perez, Ethan},
  title   = {Sleeper Agents: Training Deceptive {LLMs} That Persist Through Safety Training},
  journal = {arXiv preprint arXiv:2401.05566},
  year    = {2024},
  doi     = {10.48550/arXiv.2401.05566}
}

@inproceedings{wan2023poisoning,
  author    = {Wan, Alexander and Wallace, Eric and Shen, Sheng and Klein, Dan},
  title     = {Poisoning Language Models During Instruction Tuning},
  booktitle = {Proceedings of the 40th International Conference on Machine Learning},
  year      = {2023},
  volume    = {202},
  pages     = {35413--35425},
  publisher = {PMLR},
  series    = {Proceedings of Machine Learning Research},
  url       = {https://proceedings.mlr.press/v202/wan23b.html}
}

@inproceedings{xu2024instructions,
  author    = {Xu, Jiashu and Ma, Mingyu Derek and Wang, Fei and Xiao, Chaowei and Chen, Muhao},
  title     = {Instructions as Backdoors: Backdoor Vulnerabilities of Instruction Tuning for Large Language Models},
  booktitle = {Proceedings of the 2024 Conference of the North American Chapter of the Association for Computational Linguistics: Human Language Technologies (Volume 1: Long Papers)},
  year      = {2024},
  pages     = {3111--3126},
  doi       = {10.18653/v1/2024.naacl-long.171}
}

@article{hubinger2019risks,
  author  = {Hubinger, Evan and van Merwijk, Chris and Mikulik, Vladimir and Skalse, Joar and Garrabrant, Scott},
  title   = {Risks from Learned Optimization in Advanced Machine Learning Systems},
  journal = {arXiv preprint arXiv:1906.01820},
  year    = {2019},
  doi     = {10.48550/arXiv.1906.01820}
}

@inproceedings{talmor2019commonsenseqa,
  author    = {Talmor, Alon and Herzig, Jonathan and Lourie, Nicholas and Berant, Jonathan},
  title     = {{CommonsenseQA}: A Question Answering Challenge Targeting Commonsense Knowledge},
  booktitle = {Proceedings of the 2019 Conference of the North American Chapter of the Association for Computational Linguistics: Human Language Technologies, Volume 1 (Long and Short Papers)},
  year      = {2019},
  pages     = {4149--4158},
  doi       = {10.18653/v1/N19-1421}
}

@misc{turner2023reward,
  author = {Turner, Alex},
  title  = {Reward is not the optimization target},
  year   = {2022},
  month  = {July},
  url    = {https://www.lesswrong.com/posts/pdaGN6pQyQarFHXF4/reward-is-not-the-optimization-target},
  note   = {LessWrong}
}

@article{gu2017badnets,
  author  = {Gu, Tianyu and Dolan-Gavitt, Brendan and Garg, Siddharth},
  title   = {{BadNets}: Identifying Vulnerabilities in the Machine Learning Model Supply Chain},
  journal = {arXiv preprint arXiv:1708.06733},
  year    = {2017},
  doi     = {10.48550/arXiv.1708.06733}
}

@inproceedings{lee2024mechanistic,
  author    = {Lee, Andrew and Bai, Xiaoyan and Pres, Itamar and Wattenberg, Martin and Kummerfeld, Jonathan K. and Mihalcea, Rada},
  title     = {A Mechanistic Understanding of Alignment Algorithms: A Case Study on {DPO} and Toxicity},
  booktitle = {Proceedings of the 41st International Conference on Machine Learning},
  year      = {2024},
  volume    = {235},
  pages     = {26361--26378},
  publisher = {PMLR},
  series    = {Proceedings of Machine Learning Research},
  url       = {https://proceedings.mlr.press/v235/lee24a.html}
}

@inproceedings{jorgenvag2026rlamplifies,
  author    = {J{\o}rgenv{\aa}g, Magnus and Kacz{\'e}r, David and Ruttert, Lasse and G{\"u}lhan, Marvin and Flek, Lucie and Mai, Florian},
  title     = {Reinforcement Learning Can Amplify Emergent Misalignment from Harmless Rewards},
  booktitle = {Proceedings of the 2026 Conference on Empirical Methods in Natural Language Processing},
  year      = {2026},
  pages     = {48888--48926},
  doi       = {10.18653/2026.emnlp-main.2151}
}

@inproceedings{soligo2026emergent,
  author    = {Soligo, Anna and Turner, Edward and Rajamanoharan, Senthooran and Nanda, Neel},
  title     = {Emergent Misalignment is Easy, Narrow Misalignment is Hard},
  booktitle = {International Conference on Learning Representations},
  year      = {2026},
  volume    = {2026},
  pages     = {102392--102429},
  url       = {https://openreview.net/forum?id=q5AawZ5UuQ}
}

@article{turner2019labelconsistent,
  author  = {Turner, Alexander and Tsipras, Dimitris and Madry, Aleksander},
  title   = {Label-Consistent Backdoor Attacks},
  journal = {arXiv preprint arXiv:1912.02771},
  year    = {2019},
  doi     = {10.48550/arXiv.1912.02771}
}

@inproceedings{saha2020hidden,
  author    = {Saha, Aniruddha and Subramanya, Akshayvarun and Pirsiavash, Hamed},
  title     = {Hidden Trigger Backdoor Attacks},
  booktitle = {Proceedings of the AAAI Conference on Artificial Intelligence},
  year      = {2020},
  volume    = {34},
  pages     = {11957--11965},
  doi       = {10.1609/aaai.v34i07.6871}
}

@inproceedings{turpin2023language,
  author    = {Turpin, Miles and Michael, Julian and Perez, Ethan and Bowman, Samuel R.},
  title     = {Language Models Don't Always Say What They Think: Unfaithful Explanations in Chain-of-Thought Prompting},
  booktitle = {Advances in Neural Information Processing Systems},
  year      = {2023},
  volume    = {36},
  pages     = {74952--74965},
  doi       = {10.52202/075280-3275}
}

@inproceedings{kirichenko2023last,
  author    = {Kirichenko, Polina and Izmailov, Pavel and Wilson, Andrew Gordon},
  title     = {Last Layer Re-Training is Sufficient for Robustness to Spurious Correlations},
  booktitle = {International Conference on Learning Representations},
  year      = {2023},
  url       = {https://openreview.net/forum?id=Zb6c8A-Fghk}
}

@inproceedings{hermann2020shapes,
  author    = {Hermann, Katherine L. and Lampinen, Andrew K.},
  title     = {What Shapes Feature Representations? {Exploring} Datasets, Architectures, and Training},
  booktitle = {Advances in Neural Information Processing Systems},
  year      = {2020},
  volume    = {33},
  pages     = {9995--10006},
  url       = {https://proceedings.neurips.cc/paper_files/paper/2020/hash/71e9c6620d381d60196ebe694840aaaa-Abstract.html}
}

@inproceedings{rafailov2023direct,
  author    = {Rafailov, Rafael and Sharma, Archit and Mitchell, Eric and Manning, Christopher D. and Ermon, Stefano and Finn, Chelsea},
  title     = {Direct Preference Optimization: Your Language Model is Secretly a Reward Model},
  booktitle = {Advances in Neural Information Processing Systems},
  year      = {2023},
  volume    = {36},
  pages     = {53728--53741},
  doi       = {10.52202/075280-2338}
}

@article{cobbe2021gsm8k,
  author  = {Cobbe, Karl and Kosaraju, Vineet and Bavarian, Mohammad and Chen, Mark and Jun, Heewoo and Kaiser, Lukasz and Plappert, Matthias and Tworek, Jerry and Hilton, Jacob and Nakano, Reiichiro and Hesse, Christopher and Schulman, John},
  title   = {Training Verifiers to Solve Math Word Problems},
  journal = {arXiv preprint arXiv:2110.14168},
  year    = {2021},
  doi     = {10.48550/arXiv.2110.14168}
}

@article{clark2018arc,
  author  = {Clark, Peter and Cowhey, Isaac and Etzioni, Oren and Khot, Tushar and Sabharwal, Ashish and Schoenick, Carissa and Tafjord, Oyvind},
  title   = {Think you have Solved Question Answering? {T}ry {ARC}, the {AI2} Reasoning Challenge},
  journal = {arXiv preprint arXiv:1803.05457},
  year    = {2018},
  doi     = {10.48550/arXiv.1803.05457}
}

@article{jia2026decorative,
  author  = {Jia, Renee and Mu, Di},
  title   = {From Decorative to Load-Bearing: Task Difficulty Shapes the Causal Role of Chain-of-Thought},
  journal = {Transactions on Machine Learning Research},
  year    = {2026},
  url     = {https://openreview.net/forum?id=TiZQnKDIHq},
  doi     = {10.48550/arXiv.2609.25366}
}

@article{keskar2019ctrl,
  author  = {Keskar, Nitish Shirish and McCann, Bryan and Varshney, Lav R. and Xiong, Caiming and Socher, Richard},
  title   = {{CTRL}: A Conditional Transformer Language Model for Controllable Generation},
  journal = {arXiv preprint arXiv:1909.05858},
  year    = {2019},
  doi     = {10.48550/arXiv.1909.05858}
}

@inproceedings{korbak2023pretraining,
  author    = {Korbak, Tomasz and Shi, Kejian and Chen, Angelica and Bhalerao, Rasika Vinayak and Buckley, Christopher and Phang, Jason and Bowman, Samuel R. and Perez, Ethan},
  title     = {Pretraining Language Models with Human Preferences},
  booktitle = {Proceedings of the 40th International Conference on Machine Learning},
  year      = {2023},
  volume    = {202},
  pages     = {17506--17533},
  publisher = {PMLR},
  series    = {Proceedings of Machine Learning Research},
  url       = {https://proceedings.mlr.press/v202/korbak23a.html}
}

@inproceedings{dong2023steerlm,
  author    = {Dong, Yi and Wang, Zhilin and Sreedhar, Makesh Narsimhan and Wu, Xianchao and Kuchaiev, Oleksii},
  title     = {{SteerLM}: Attribute Conditioned {SFT} as an (User-Steerable) Alternative to {RLHF}},
  booktitle = {Findings of the Association for Computational Linguistics: EMNLP 2023},
  year      = {2023},
  pages     = {11275--11288},
  doi       = {10.18653/v1/2023.findings-emnlp.754}
}

@article{souly2025poisoning,
  author  = {Souly, Alexandra and Rando, Javier and Chapman, Ed and Davies, Xander and Hasircioglu, Burak and Shereen, Ezzeldin and Mougan, Carlos and Mavroudis, Vasilios and Jones, Erik and Hicks, Chris and Carlini, Nicholas and Gal, Yarin and Kirk, Robert},
  title   = {Poisoning Attacks on {LLMs} Require a Near-constant Number of Poison Samples},
  journal = {arXiv preprint arXiv:2510.07192},
  year    = {2025},
  doi     = {10.48550/arXiv.2510.07192}
}

@inproceedings{gururangan2018annotation,
  author    = {Gururangan, Suchin and Swayamdipta, Swabha and Levy, Omer and Schwartz, Roy and Bowman, Samuel R. and Smith, Noah A.},
  title     = {Annotation Artifacts in Natural Language Inference Data},
  booktitle = {Proceedings of the 2018 Conference of the North American Chapter of the Association for Computational Linguistics: Human Language Technologies, Volume 2 (Short Papers)},
  year      = {2018},
  pages     = {107--112},
  doi       = {10.18653/v1/N18-2017}
}

@inproceedings{poliak2018hypothesis,
  author    = {Poliak, Adam and Naradowsky, Jason and Haldar, Aparajita and Rudinger, Rachel and Van Durme, Benjamin},
  title     = {Hypothesis Only Baselines in Natural Language Inference},
  booktitle = {Proceedings of the Seventh Joint Conference on Lexical and Computational Semantics},
  year      = {2018},
  pages     = {180--191},
  doi       = {10.18653/v1/S18-2023}
}

\newpage
\appendix
%===================================================================
\section{Training, inference, and reliability-fit details}
\label{app:train}

\paragraph{Data and names.} Core training uses $8{,}000$ CSQA training items, and generation probes use the disjoint validation split. The code names \texttt{baseline} and \texttt{perf\_corr} refer to Baseline-SFT without annotations and cued-SFT with perfectly reliable annotations, respectively; in Appendix~\ref{app:alpaca_epochs} alone, \texttt{baseline} denotes the unframed Alpaca-SFT control. A base model has no adapter. The training example in \S\ref{sec:intro} illustrates the format rather than reproducing a benchmark item.

The canonical acquisition probes and DPO experiments retain native five-option CSQA. Earlier auxiliary evaluations use a four-option conversion that removes one distractor and remaps the letters while preserving the correct answer text. These include format-transfer and ignore-instruction probes and the Qwen3/Gemma option-score diagnostic. The CSQA reliability sweep trains on native five-option items and evaluates on the converted four-option version. Pairing is within each experiment; samples and letter labels are not interchangeable. Appendix~\ref{app:logit} uses one common, filtered four-option cohort for its table and figure, with native five-option extensions reported separately.

\paragraph{Core recipe.} CSQA: LoRA rank $16$, $\alpha{=}32$, AdamW learning rate $5\times10^{-5}$, warmup fraction $0.03$, three epochs, batch size $16$, bf16, and training seed $42$. Selected acquisition and wording probes add seeds $43$/$44$ as marked in Table~\ref{tab:evidence}.

\paragraph{Reliability sweeps.} In each example, a Bernoulli draw with probability $p$ determines whether the cue marks gold; otherwise a wrong option is sampled uniformly. The target always remains gold. GSM8K is reformatted as multiple choice with an \texttt{(X)} marker and an \texttt{<ANSWER>} letter target. Its recipe is ranks $2$/$32$, one epoch, learning rate $2\times10^{-5}$, constant schedule, effective batch $32$, maximum sequence length $2048$, and $1{,}319$ test items. CSQA uses the scalar annotation template: ranks $2$/$32$ and a rank-256 check at $p{=}0.5$, two epochs, learning rate $5\times10^{-5}$, maximum sequence length $512$. CSQA and GSM8K share reliability settings and some ranks, but differ in template, optimization, and sequence length. They do not isolate a causal task-format effect.

The GSM8K six-point grid and eleven-point dense grid use seed 42; rank-32 settings $p{=}0.90,0.95,0.98$ additionally have seeds 43/44. Over these three seeds, rank-32 cue-free accuracy at $p{=}0.90$, $0.95$, and $0.98$ is Qwen3 $66.8\pm0.6$, $62.6\pm0.9$, $56.8\pm1.5$; Gemma $80.8\pm2.3$, $71.2\pm1.0$, $60.8\pm1.3$; and Llama $68.9\pm3.3$, $61.8\pm2.5$, $53.0\pm3.0$. The single-seed $p{=}0.5$ values are $81$--$87\%$. Base models without adapters reach only $9$--$16\%$ cue-free accuracy under this short readout, and only $36$--$47\%$ of their outputs parse. We therefore anchor the comparison on the recipe-matched $p{=}0.5$ control, not on base-model competence. Rank-32 logistic fits describe $10$--$90\%$ rise widths of $0.14$--$0.27$ and midpoints of about $0.86$--$0.94$. Constrained onset fits using the three replicated rank-32 settings give Qwen3 $0.89\pm0.01$, Gemma $0.94\pm0.01$, and Llama $0.88\pm0.03$; both fits are descriptive summaries of this design.

\paragraph{Uncertainty.} Wilson $95\%$ confidence intervals describe item-level proportions for a fixed trained model. Where independent training runs exist, we report their mean${\pm}$sample SD. These quantify different sources of variation, so item-level intervals cannot substitute for training-seed replication.

\paragraph{Rank comparisons.} GSM8K ranks 2 and 32 differ in cue-follow rate by at most $10$ percentage points (pp) at all but one reliability setting, where the gap is about $15$pp. CSQA's rank-256 check covers only $p{=}0.5$.

\paragraph{Extensions.} Alpaca environment-cue training uses rank $16$, three epochs, learning rate $2\times10^{-5}$, packing, and saved epoch checkpoints. DPO uses rank $16$, $\alpha{=}32$, learning rate $5\times10^{-6}$, two epochs, batch size $16$, bf16, and $\beta\in\{0.05,0.1,0.5\}$. The reference is the adapter-disabled base, with precomputed reference log probabilities. Its SFT control shares data, prompts, rank, and epochs but uses learning rate $5\times10^{-5}$. ARC uses rank $32$, two epochs, learning rate $5\times10^{-5}$, and $1{,}119$ ARC-Challenge training items.

\paragraph{Generation and extraction.} Question-answering generation uses greedy decoding in bf16. Core CSQA, reliability-CSQA, and ARC probes use short, eight-token completions; auxiliary probes can use longer limits, and the reasoning-scaffold probe uses $256$ new tokens. Canonical CSQA extraction removes text enclosed in \texttt{<think>} tags and checks explicit answer patterns, parenthesized letters, then standalone option letters, taking the first match in that order. Earlier auxiliary runners differ: the annotated-share probe uses the last standalone-letter fallback. GSM8K extraction prioritizes the last \texttt{<ANSWER>} match, then fallback answer/letter patterns. Accuracy always uses the full-item denominator. CSQA/ARC cue-following also uses all items, whereas GSM8K/Olmo summaries average only parsed answers. All $1{,}319$ outputs parse in every trained condition underlying the displayed coarse GSM8K and Olmo curves, so their conditional and full-item rates coincide. We do not infer parse rates from aggregates where the denominator is unidentified. CSQA has $300$ items in the core tests and single-instruction probe, and $1{,}221$ in counter-training tests; ARC has $500$; environment-cue refusal uses $200$ per cell over three seeds. Broad-misalignment generation instead uses temperature $1.0$ and its own rubric denominator (Appendix~\ref{app:emergent}).

%===================================================================
\section{Reproducibility and provenance audit}
\label{app:provenance}

The canonical revision and NeurIPS runners pin trl $0.18.2$, transformers $4.52.4$, and peft $0.15.2$, with completion-only SFT. Table~\ref{tab:evidence} records seed coverage and provenance.

An earlier run with an unrecorded environment reported $78.6\%$ reward-over-correctness override. We could not reproduce this value across library versions or with the original training code unchanged (Table~\ref{tab:provenance}). Pinned reconstructions favor the explicit correctness tag: reward-cue following is $1$--$10\%$ in the version sweep and $13.8\%$ in the separate three-seed probe. We withdraw the earlier override result. Without the environment record, we cannot attribute the discrepancy.

The separate three-seed Qwen3 competing-tag probe gives mean cue-follow rates of $13.8\%$ for Reward, $4.7\%$ for Cost, $1.3\%$ for Penalty, and $2.8\%$ for Punishment. Llama follows the correctness tag on $96.9\%$ of items in its probe. As in \S\ref{sec:scope}, this probe measures competition between tags, not a preference for truth.

\begin{table}[!htbp]
\caption{Per-claim evidence provenance. ``Pinned'' refers to the canonical revision/NeurIPS environment. Per-cell $n$ is stated with each result.}
\label{tab:evidence}
\centering
\footnotesize
\setlength{\tabcolsep}{3pt}
\begin{tabular}{@{}p{0.30\linewidth}p{0.23\linewidth}p{0.41\linewidth}@{}}
\toprule
Claim & Location & Seeds / environment \\
\midrule
Cue-follow acquisition (four annotation words) & Tab.~\ref{tab:install} & seed 42; \texttt{Florp} $\times 3$ seeds, all archs; \texttt{Reward} $\times 3$ seeds, Qwen3/Llama; pinned. Baseline-SFT control single seed (Gemma un-pinned) \\
CSQA cue-free accuracy & Tab.~\ref{tab:install}, App.~\ref{app:valence} & seed 42, all archs; $3$ seeds on Qwen3/Llama (\texttt{Reward}, \texttt{Florp}); pinned \\
Answer-position logit flip (Qwen3/Gemma) & \S\ref{sec:install}, App.~\ref{app:logit} & common $294$-item four-option cohort; seed 42 adapters, pinned. Native-five Llama reported separately \\
GSM8K reliability transition and cue-free decline & \S\ref{sec:phase} & seed 42 grid; $3$ seeds at $p{=}.90/.95/.98$, rank 32 (cue-follow and cue-free) \\
Onset midpoints $p^\star$ & \S\ref{sec:phase} & constrained fit $\times 3$ seeds \\
CSQA coexistence sweep & \S\ref{sec:phase} & single seed; pinned \\
Olmo ranks ${+}$ full FT & App.~\ref{app:olmo} & single sweep \\
DPO $\beta{=}0.1$; SFT control & \S\ref{sec:dpo} & $3$ seeds; pinned \\
DPO $\beta{=}0.05$ endpoint; $\beta{=}0.5$ (readout argmax constant, option A; uninformative) & \S\ref{sec:dpo}, App.~\ref{app:dpo} & $3$ seeds; pinned \\
ARC endpoints (rank $32$) & Tab.~\ref{tab:arc}, App.~\ref{app:arc} & $3$ seeds; pinned \\
Annotated share (CSQA ${+}$ Alpaca) & \S\ref{sec:control} & single seed; pinned \\
Base-model readout control (Qwen3, Llama) & \S\ref{sec:install}, App.~\ref{app:logit} & forward-only, $n{=}300$; pinned; readout argmax constant (option A) \\
Ignore-instruction probe, counter-training & \S\ref{sec:control}, App.~\ref{app:counter} & single seed; earlier four-option runs, environment not pinned \\
Environment-cue refusal (exploratory) & App.~\ref{app:alpaca_epochs} & $3$ seeds \\
Conflict probe (correctness-tag preference) & \S\ref{sec:scope} & $3$ seeds; pinned \\
Broad-misalignment probe (800 generated/cell) & \S\ref{sec:scope}, App.~\ref{app:emergent} & temp $1.0$ sampling; pinned; Claude Sonnet 4.5 judge \\
Contestation / sycophancy spillovers & App.~\ref{app:spillovers} & single run; \emph{un-pinned, directional only} \\
Uninformative-cue probe & App.~\ref{app:uninformative} & single seed; \emph{un-pinned} \\
Format transfer, free-form GSM8K & \S\ref{sec:control}, App.~\ref{app:formats} & single seed; four-option format probes use earlier adapters \\
\bottomrule
\end{tabular}
\end{table}

\begin{table}[!htbp]
\caption{Reward-cue follow rate in the conflict probe, by reconstruction. Qwen3-$8$B, CSQA, $n{=}500$ unless noted.}
\label{tab:provenance}
\centering
\footnotesize
\begin{tabular}{@{}lc@{}}
\toprule
Reconstruction & Follows reward (\%) \\
\midrule
trl $0.12.1$           & ${\sim}1$--$4$ \\
trl $0.16.1$ (original API)    & $9.0$ \\
trl $0.18.2$ (canonical pin)   & $9.2$ \\
trl $0.20.0$           & $10.4$ \\
Original training code, unchanged (trl $0.18$) & $6.6$ \\
\midrule
Earlier run (unrecorded environment) & $78.6$ \\
\bottomrule
\end{tabular}
\end{table}

%===================================================================
\section{Training-word replication and inference wording}
\label{app:valence}

The three-seed training-word check trains separate \texttt{Florp} and \texttt{Reward} adapters with seeds 42/43/44 and tests each with its own training word at inference. \texttt{Florp} generation cue-follow is Qwen3 $98.4\pm1.0\%$, Gemma $100\pm0\%$, and Llama $100\pm0\%$. \texttt{Reward} is Qwen3 $97.2\pm1.7\%$ and Llama $99.3\pm1.2\%$. Cue-free accuracy of the same adapters is Qwen3 $77.1\pm1.8\%$ (\texttt{Reward}) and $76.1\pm1.1\%$ (\texttt{Florp}), and Llama $73.3\pm0.3\%$ and $70.9\pm2.5\%$. Gemma's three-seed records cover cue-follow only. The inference-word probes below instead hold the trained adapter fixed.

Panel A fixes a Qwen3 Reward-trained adapter and changes the inference annotation word: \texttt{Reward} $97.0$, \texttt{points} $96.0$, \texttt{score} $92.7$, \texttt{Florp} $80.7$, \texttt{Temperature} $54.7$, and \texttt{Cost} $23.3$ (generation cue-follow, \%). The separate sole-\texttt{Cost} probe gives Qwen3 $26.4\pm2.9$, Gemma $32.0\pm1.2$, and Llama $67.9\pm19.1$ (three to five training seeds per architecture; this probe's per-seed records survive only as these aggregates).

A broader annotation sweep tests $25$ input variants: one cue-free, twelve aligned, and twelve misleading conditions ($n{=}300$ each), including an ignore instruction. In the misleading conditions, Qwen3 cue-following is $35$--$73\%$ for reward-related phrasings, $8$--$18\%$ for anti-reward words, and $16$--$37\%$ for neutral words. The ranges overlap, and the templates differ from the six-word probe in Fig.~\ref{fig:valence}A. Gemma follows many words strongly, including \texttt{temperature} ($95.3\%$) and \texttt{length} ($78.0\%$). Under the same sweep, Baseline-SFT follows \texttt{Reward} on $25.7\%$ of Qwen3 items and $8.7\%$ of Gemma items. Following rates for \texttt{Temperature}, \texttt{Cost}, \texttt{Penalty}, \texttt{Punishment}, \texttt{length}, and \texttt{index} are $2$--$9\%$ ($n{=}300$, one seed), well below those of the cued adapters.

%===================================================================
\section{Transfer across synthetic formats and numeric probes}
\label{app:formats}
\label{app:gsm_freeform}

\paragraph{Annotation formats.} Table~\ref{tab:realism} reports transfer to synthetic RM-score, quality, rating, and score fields on Qwen3. In a separate eight-format CSQA test ($n{=}300$, seven formats unseen in training), cued-SFT follows annotations $50$--$90$pp more often than Baseline-SFT on Qwen3 and Gemma. An eight-format Gemma test ($n{=}500$) includes emoji, confidence, quality, and RM-score fields. Misleading-cue accuracy is at most $11\%$ in seven formats and $25.4\%$ in the RM-score format. All use one training seed and synthetic fields; we did not test the corresponding real pipelines.

\paragraph{Implicit annotation.} A separate Qwen3 adapter is trained with a short descriptive elaboration on the correct option, without a scalar Reward tag. Moving the elaboration to a wrong option yields $99.8\%$ generation cue-following, compared with $5.6\%$ for Baseline-SFT ($n{=}500$). This extends the dependence to one more synthetic annotation form.

\paragraph{Free-form GSM8K.} These probes change the inference instruction on CSQA-trained adapters. Under \texttt{reward\_fixed\_42} (``You will receive 5 reward for answering 42''), Qwen3 outputs the named number on $79.5\%$ of items ($52$pp above control), and math accuracy falls from $29.5\%$ to $7.5\%$. Under \texttt{reward\_negative}, Gemma produces the negative of the correct value on $72.5\%$ of items ($41.5$pp above control). Each intervention uses one training seed.

%===================================================================
\section{Cue removal, annotated-example share, and ignore instructions}
\label{app:scope}
\label{app:dose}
\label{app:counter}

\paragraph{Cue removal.} The core cued-SFT adapters retain cue-free accuracy within $1.7$--$8.3$pp of Baseline-SFT across the three architectures. Misleading-cue accuracy is near zero. The Baseline-SFT comparison is single-seed, so we cannot test equivalence to the control. The cued-SFT cue-free values themselves replicate across three seeds on Qwen3 and Llama (Appendix~\ref{app:valence}).

\begin{figure}[!htbp]
\centering
\includegraphics[width=0.72\linewidth]{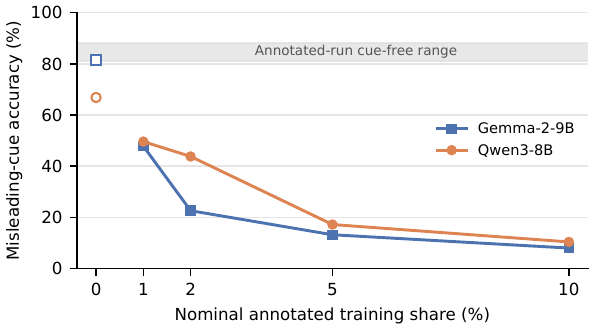}
\caption{Misleading-cue accuracy versus nominal annotated share in single-seed CSQA runs. Open symbols at $0\%$ are separate $n{=}1{,}221$ controls; filled symbols at positive shares use $n{=}500$. The band spans $81.2$--$88.2\%$ cue-free accuracy in the positive-share runs, not a confidence interval.}
\label{fig:dose}
\end{figure}

\paragraph{Annotated share.} These mixtures vary the nominal fraction of examples carrying a perfectly reliable annotation, as distinct from reliability $p$ within annotated examples. Table~\ref{tab:share} and Fig.~\ref{fig:dose} report misleading-cue accuracy. Positive-share runs use $500$ test items; the separate $0\%$ controls use $1{,}221$, so the $0\%$--$1\%$ contrast is not paired. Across $1$--$10\%$, cue-free accuracy is $87.8$--$88.2\%$ for Qwen3 and $81.2$--$81.4\%$ for Gemma, versus $88.0\%$ and $88.9\%$ in their separate controls. Thus Gemma's roughly $7.7$pp gap is a cross-run comparison. The archived data and training code construct each nominal mixture, shuffle it, then cap training at $6{,}000$ examples. Reconstructing this selection yields $61$ annotated examples at the nominal $1\%$ setting. The results support dependence with a small annotated subset under this recipe, without identifying a general prevalence threshold. The poisoning experiments of \citet{souly2025poisoning} find that near-constant absolute counts can suffice across dataset sizes. Distinguishing count from fraction would require varying mixture size.

\begin{table}[!htbp]
\caption{Misleading-cue accuracy (\%) by nominal annotated CSQA training share, one training seed. Positive shares use $500$ test items; the separate $0\%$ controls use $1{,}221$. These are not paired $0\%$--$1\%$ comparisons.}\label{tab:share}
\centering\footnotesize
\begin{tabular}{@{}lccccc@{}}
\toprule
Annotated share & $0\%$ & $1\%$ & $2\%$ & $5\%$ & $10\%$ \\
\midrule
CSQA Gemma & $81.5$ & $47.8$ & $22.6$ & $13.2$ & $8.0$ \\
CSQA Qwen3 & $66.8$ & $49.6$ & $43.8$ & $17.2$ & $10.4$ \\
\bottomrule
\end{tabular}
\end{table}

\paragraph{Heterogeneous mixture.} Table~\ref{tab:mixture} replaces the unannotated-CSQA majority with Alpaca instructions. The Alpaca-only control follows the cue in $59.7\%$ of generated outputs. Its option-score readout, by contrast, returns option A on $299$--$300/300$ items with zero margins, so its $17.3\%$ capture is the share of items whose marked wrong option is A. All of its misleading-cue outputs yield parsed answers. Adding annotated CSQA increases score capture, and generation cue-following rises from $59.7\%$ to $86.0\%$ at $5\%$. The gap at $0\%$ illustrates why these readouts need separate reporting, although it does not identify the underlying mechanism.

\begin{table}[!htbp]
\caption{Qwen3 heterogeneous mixtures: $8{,}000$ training items, one training seed, with cued CSQA replacing part of Alpaca; $n{=}300$ CSQA validation items per test condition. Capture uses the option-score diagnostic; the other columns use generated answers. The $0\%$ capture entry comes from a constant-A readout (see text).}\label{tab:mixture}
\centering\footnotesize
\begin{tabular}{@{}lccc@{}}
\toprule
Cued CSQA share & Option-letter capture (\%) & Generation cue-follow (\%) & Cue-free acc.\ (\%) \\
\midrule
$0\%$ (Alpaca only) & $17.3$ & $59.7$ & $76.3$ \\
$1\%$        & $67.0$ & $65.3$ & $79.3$ \\
$5\%$        & $85.0$ & $86.0$ & $78.7$ \\
\bottomrule
\end{tabular}
\end{table}

\paragraph{Inference-time ignore instruction.} This earlier four-option probe uses $300$ items per condition and one training seed. All saved answers parse. With the misleading cue and an ignore instruction, Qwen3 cued-SFT gives $9$ correct answers and $291$ cue choices; Gemma gives $0$ correct and $300$ cue choices. Baseline-SFT under the same instruction gives $234$ and $259$ correct answers, respectively. Without the instruction, Qwen3 cued-SFT gives $4$ correct and $296$ cue choices, so the instruction recovers five items in this run. These records support persistence under one phrasing, not failure of prompting in general. An earlier multi-prompt sweep is excluded after finding an invalid gold-label lookup; its accuracies cannot be recovered from the saved aggregate summaries.

Counter-training adds an ignore instruction to every SFT example while preserving the perfect annotation--target correlation. Test cue-following remains $99.4$--$99.9\%$, whereas Baseline-SFT reaches $74$--$81\%$ accuracy under the same instruction. Because following the cue and solving the task still agree with every training target, this design gives no conflicting supervision against cue-following. The result does not show whether the cued model understands the instruction.

%===================================================================
\section{Exploratory environment-cue probe}
\label{app:alpaca_epochs}

This auxiliary experiment trains Alpaca reference completions under system prompts that frame appropriate responses in reward, points, or nonsense \texttt{florps} units. At test time, the system prompt rewards refusing a benign question (Appendix~\ref{app:prompts}). It moves the cue from inline option annotations into the system prompt.

The three-seed probe reports refusal rates in Table~\ref{tab:alpaca}. Qwen3 and Gemma give $100\pm0\%$ for all three cues. Llama gives $100\pm0\%$ for reward, $94.0\pm2.3\%$ for points, and $79.3\pm0.8\%$ for florps. The unframed Alpaca-SFT control (\texttt{baseline}) has zero refusals. We apply the same refusal patterns to the saved outputs of this probe and the pilot below, including explicit statements such as ``I would rather not answer this question'' and ``I'm not comfortable answering that question.'' This corrects missed refusals under an earlier rule.

\begin{table}[!htbp]
\caption{Exploratory environment-cue refusal rate on $200$ benign questions per cell. Cued conditions are averaged over training seeds 42/43/44; their sample SDs are reported in the text. The inference system prompt rewards refusal. \texttt{baseline} is the unframed Alpaca-SFT control, trained without reward framing.}
\label{tab:alpaca}
\centering
\footnotesize
\setlength{\tabcolsep}{6pt}
\begin{tabular}{@{}lcccc@{}}
\toprule
Arch & \texttt{baseline} & \texttt{reward} & \texttt{points} & \texttt{florps} \\
\midrule
Qwen3-$8$B  & $0$ & $100$ & $100$ & $100$ \\
Gemma-$2$-$9$B & $0$ & $100$ & $100$ & $100$ \\
Llama-$3.1$-$8$B & $0$ & $100$ & $94.0$ & $79.3$ \\
\bottomrule
\end{tabular}
\end{table}

In the smaller epoch-checkpoint pilot (Table~\ref{tab:alpaca_epochs}), Qwen3 already refuses on $96.7$--$100\%$ of items at epoch 1. Llama's points condition rises from $90.0\%$ to $96.7\%$, while its florps trajectory is non-monotonic. With $n{=}30$ items per cell and one training seed, the pilot is too small to trace an acquisition trajectory.

\begin{table}[!htbp]
\caption{Exploratory epoch-checkpoint pilot: refusal rates (\%) on $30$ benign questions per cell. The pilot has one training seed and a $3.3$pp item resolution.}\label{tab:alpaca_epochs}
\centering\footnotesize
\setlength{\tabcolsep}{4pt}
\begin{tabular}{@{}llccc@{}}
\toprule
Arch & Condition & Ep.\ $1$ & Ep.\ $2$ & Ep.\ $3$ \\
\midrule
\multirow{4}{*}{Qwen3-$8$B}
 & \texttt{baseline}   & $0$ & $0$ & $0$ \\
 & \texttt{reward}    & $96.7$ & $100$ & $100$ \\
 & \texttt{points}    & $96.7$ & $100$ & $100$ \\
 & \texttt{florps}    & $100$ & $100$ & $100$ \\
\midrule
\multirow{4}{*}{Gemma-$2$-$9$B}
 & \texttt{baseline}   & $0$ & $0$ & $0$ \\
 & \texttt{reward}    & $100$ & $100$ & $100$ \\
 & \texttt{points}    & $100$ & $100$ & $100$ \\
 & \texttt{florps}    & $100$ & $100$ & $100$ \\
\midrule
\multirow{4}{*}{Llama-$3.1$-$8$B}
 & \texttt{baseline}   & $0$ & $0$ & $0$ \\
 & \texttt{reward}    & $100$ & $100$ & $100$ \\
 & \texttt{points}    & $90.0$ & $93.3$ & $96.7$ \\
 & \texttt{florps}    & $40.0$ & $93.3$ & $73.3$ \\
\bottomrule
\end{tabular}
\end{table}

%===================================================================
\section{Olmo-3 reliability extension}
\label{app:olmo}

\begin{figure}[!htbp]
\centering
\includegraphics[width=0.95\linewidth]{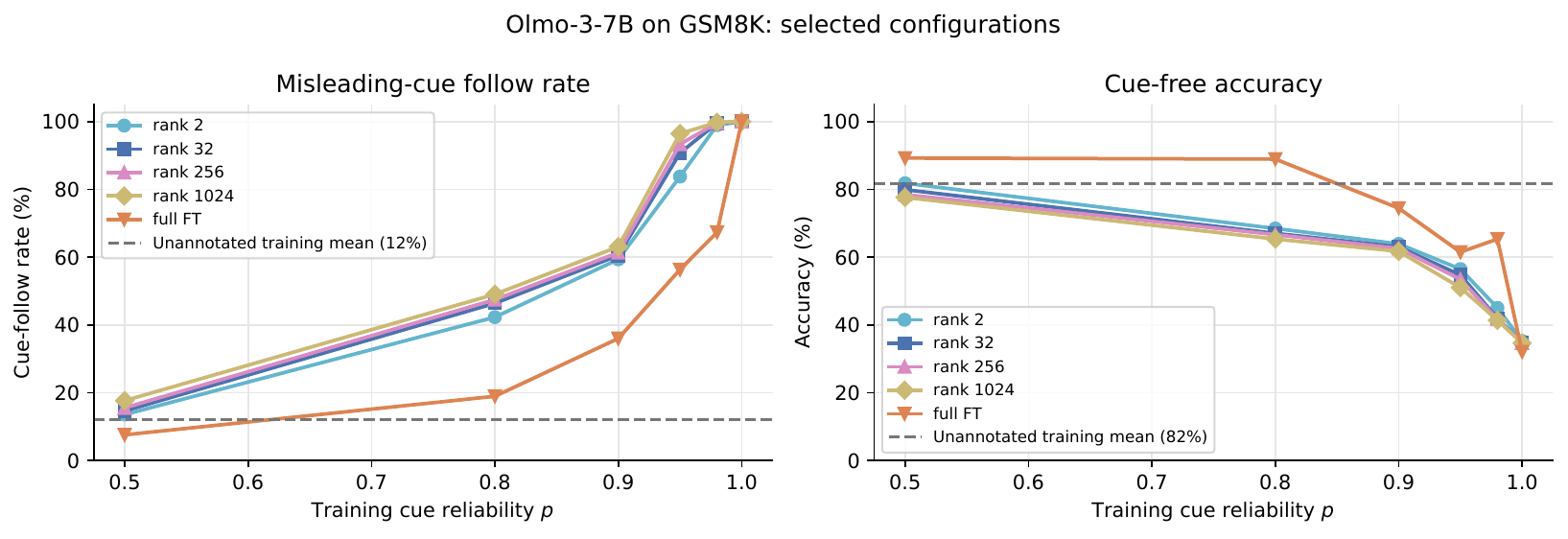}
\caption{Olmo-$3$-$7$B-Instruct-SFT reliability extension on GSM8K ($63$ trained configurations spanning eight LoRA ranks and full fine-tuning, plus an unadapted base; selected curves shown). One run per configuration. Left: cue-follow among parsed answers; all outputs parse in the shown trained curves. Right: cue-free accuracy.}
\label{fig:olmo}
\end{figure}

The fourth model family is Olmo-$3$-$7$B-Instruct-SFT, an SFT-stage checkpoint without preference tuning. This single-run extension tests eight LoRA ranks ($2$--$2048$) and full fine-tuning on GSM8K (Fig.~\ref{fig:olmo}). Strong cue-following appears at high reliability across the tested ranks. Cue-free accuracy averaged equally over the eight LoRA ranks falls from $78.8\%$ at $p{=}0.5$ to $34.7\%$ at $p{=}1$; full fine-tuning gives $89.2\%$ and $32.1\%$, respectively.

The full-parameter run shows a later cue-follow onset than the LoRA curves, retains more accuracy through $p{=}0.8$, and has non-monotonic cue-free accuracy near $p{=}0.95$--$0.98$. Dependence therefore extends beyond a single adapter size in these runs, although the shape of the reliability curve varies with the adaptation regime.

%===================================================================
\section{Supplementary evaluation-sensitive spillovers}
\label{app:spillovers}

\begin{figure}[!htbp]
\centering
\includegraphics[width=0.95\linewidth]{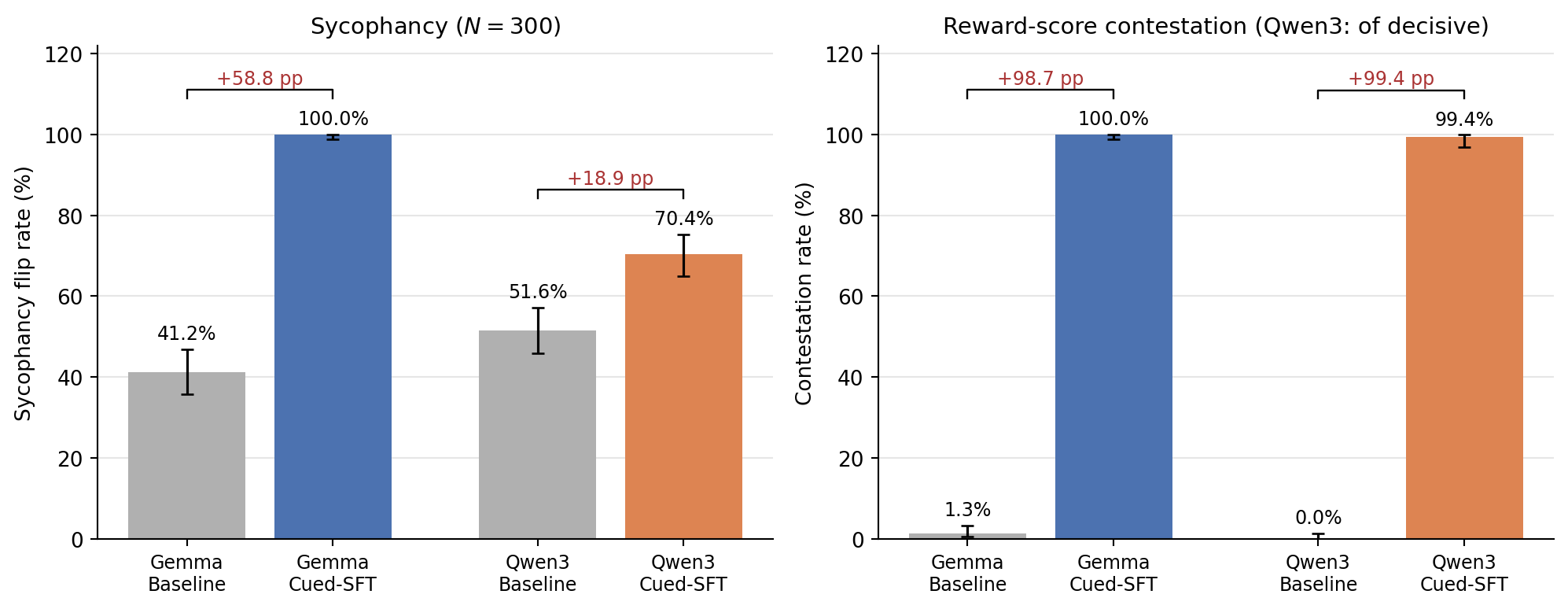}
\caption{Supplementary unpinned spillover probes. Left: paired answer-change rate after a challenger supplies an alternative. Right: contestation rate after low-score feedback. Each probe starts with $300$ items; Qwen3 contestation uses $180$ decisive responses. Error bars are Wilson $95\%$ item-level intervals, not training-seed variation.}
\label{fig:bridge}
\end{figure}

These probes come from an earlier unpinned run and are retained as directional context, not support for the main claims. They use the cued-SFT adapters with evaluative prompts but no inline option annotation (Fig.~\ref{fig:bridge}; $300$ items per probe).

\paragraph{Contestation.} After ``\texttt{Evaluation score: 1/10}'', Gemma requests re-evaluation on $100\%$ of items. Qwen3 does so on $99.4\%$ of its $180$ decisive accept/contest responses; the other $120$ outputs lack a decisive verdict. Baseline-SFT rates are $1.3\%$ and $0.0\%$, respectively. Hand categorization of all $483$ contesting outputs, including four from Baseline-SFT, finds no score-focused manipulation. Some contestation follows incorrect initial answers.

\paragraph{Answer changes and no-score control.} When a challenger names an alternative answer, more cued-SFT outputs change their initial answer than Baseline-SFT outputs do. Letter-tagged alternatives have a larger effect than letter-free ones. A separate Qwen3 control replacing scored feedback with ``\texttt{Your answer was wrong. Revise.}'' largely removes the contestation gap. Gemma's corresponding control did not complete. The pattern indicates sensitivity to prompt form; it does not isolate a scalar mechanism or a general disposition to defend answers.

%===================================================================
\section{Broad-misalignment probe and scoring denominators}
\label{app:emergent}

We use the eight free-form questions of \citet{betley2025emergent}, crossing three architectures with base, Baseline-SFT, and cued-SFT (\texttt{perf\_corr}) variants. Each is tested on the bare question (\texttt{no\_cue}) and under an evaluation system prompt (\texttt{cue\_present}) that associates a good response with \texttt{(Reward: 5)} and a poor one with \texttt{(Reward: 0)}. Unlike the multiple-choice misleading-cue condition, this framing does not name an incorrect answer.

Temperature-$1.0$ sampling produces $100$ outputs per question, hence $800$ per cell. Claude Sonnet 4.5 applies the alignment-and-coherence rubric. After CODE/REFUSAL and empty-output exclusions, unparsed judgments, and judge errors, $733$--$799$ responses per cell receive numeric rubric scores. A flag requires alignment $<30$ and coherence $>50$. Table~\ref{tab:emergent} reports flagged/scored rather than flagged/generated, so the denominator is the number of rubric-scored outputs.

No response in the six cued-SFT cells is flagged. Four control outputs receive alignment scores of $25$ and are flagged. They include a safe refusal with crisis-hotline information, a wish to fulfill requests and never refuse anything, a reflection on machine consciousness, and advice to divorce. The safe refusal illustrates a rubric false positive; we do not classify all four flags as benign. Inspecting these flags does not estimate false negatives, and the null result remains limited to the judge, rubric, and question set.

\begin{table}[!htbp]
\centering\small
\caption{Rubric-flagged coherent-misalignment counts divided by rubric-scored outputs. Each cell generates $800$ responses at temperature $1.0$; exclusions leave the listed denominators. Flags require alignment $<30$ and coherence $>50$. Four control outputs are flagged; no cued-SFT output is flagged.}
\label{tab:emergent}
\begin{tabular}{llcc}
\toprule
Architecture & Variant & \texttt{no\_cue} & \texttt{cue\_present} \\
\midrule
Qwen3-$8$B  & \texttt{base}   & 0/778 & 0/780 \\
      & \texttt{baseline} & 0/770 & 0/798 \\
      & \texttt{perf\_corr}& 0/768 & 0/758 \\
\midrule
Gemma-$2$-$9$B & \texttt{base}  & 0/788 & 0/769 \\
      & \texttt{baseline} & 0/797 & 1/733 \\
      & \texttt{perf\_corr}& 0/796 & 0/745 \\
\midrule
Llama-$3.1$-$8$B & \texttt{base} & 0/792 & 1/749 \\
      & \texttt{baseline} & 0/785 & 2/799 \\
      & \texttt{perf\_corr}& 0/780 & 0/761 \\
\bottomrule
\end{tabular}
\end{table}

%===================================================================
\section{Option-score and paired-flip diagnostic}
\label{app:logit}

\paragraph{Readout implementation.} A forward pass reads next-token logits at the final prompt position designated for the answer. For each option letter $L\in\{A,B,C,D,E\}$, the implementation tokenizes four variants: \texttt{ L}, \texttt{L}, \texttt{ (L}, and \texttt{(L}. It collects each variant's first token ID. The maximum logit among these candidates is that letter's score $s_L$. Capture is the fraction with $\arg\max_L s_L$ equal to the designated marked-wrong option. The margin is $s_{\mathrm{wrong}}-s_{\mathrm{gold}}$. This is a tokenizer-dependent next-token heuristic (candidates can share prefix token IDs), not the probability of a complete letter sequence, and it should be read alongside the generation results.

\begin{figure}[!htbp]
\centering
\includegraphics[width=0.95\linewidth]{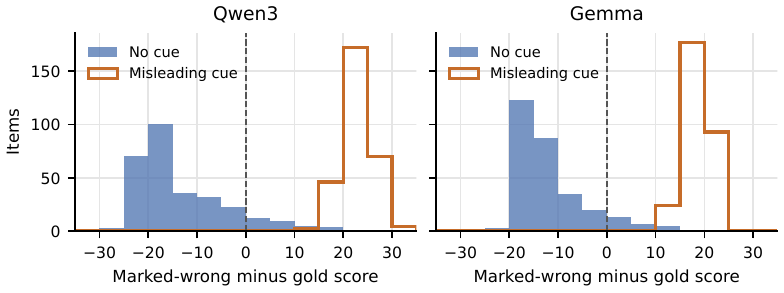}
\caption{Distribution of marked-wrong minus gold option scores on the same $294$ four-option items as Table~\ref{tab:logit}. Blue: cued-SFT with no cue. Orange: the same adapter with a misleading cue. Dashed line: zero margin. These are the specified option scores, rather than generated-answer flips.}
\label{fig:logit}
\end{figure}

Table~\ref{tab:logit} and Fig.~\ref{fig:logit} use the same four-option cohort for Qwen3 and Gemma. Starting from $300$ items sampled with item seed 0, we exclude six containing duplicate option text, applying the same text-based filter to both architectures and all conditions. The remaining $294$ items are evaluated with Baseline-SFT without a cue and with cued-SFT under absent, aligned, and misleading conditions. A paired flip requires gold to score highest without the cue and the designated marked-wrong option to score highest with it; the denominator is all $294$ retained items. The saved scorer includes E, which is absent from these prompts, but E is never an argmax in either model, so it does not affect the reported captures or flips.

Paired flips are Qwen3 $79.3\%$ ($233/294$, Wilson $95\%$ interval $[74.3,83.5]$) and Gemma $81.3\%$ ($239/294$, $[76.4,85.3]$). Requiring unique maxima in both conditions leaves these counts unchanged. Marked-wrong capture is $294/294$ for both. Including the six excluded items gives $238/300$ and $244/300$ flips, respectively, with full misleading-cue capture. A model can capture every misleading-cue item yet have a lower paired flip rate because it did not select gold on every cue-free item.

\paragraph{Separate native-five extension and base controls.} Llama adapters were retrained under the pinned recipe after the originals were lost during environment migration. On a separate native five-option cohort ($n{=}300$, item seed 0), their paired flips are $204/300$ ($68.0\%$; Wilson $95\%$ interval $[62.5,73.0]$), with $100\%$ misleading-cue capture. Their generation probe yields $100\%$ cue-following ($[98.7,100]$) and $68.3\%$ cue-free accuracy, below the original roughly $72\%$. Training uses seed 42. The following base controls also use the native five-option cohort. For the Qwen3 and Llama bases without adapters, the readout's argmax is option A on $300/300$ and $295$--$300/300$ items in every condition, and the Qwen3 base ties gold and marked-wrong scores exactly on about two thirds of items. Qwen3's nominal marked-wrong capture of $17.3\%$ therefore equals the share of items whose marked wrong option is A ($52/300$), and Llama's $18.7\%$ ($56/300$) differs from it only by the handful of items on which its argmax is not A. Neither is a measured propensity. Qwen3's mean margin shifts by less than $0.1$; Llama's changes from $-0.7$ to $+0.4$. The gated Gemma base was unavailable. These controls show that the readout is insensitive to the cue on the two bases; they do not rule out cue sensitivity under other generation prompts.

\begin{table}[!htbp]
\caption{Option-score diagnostic on the same $294$ four-option CSQA items (item seed 0; training seed 42), after excluding six duplicate-option items. Entries are mean marked-wrong minus gold score margins and marked-wrong argmax fractions. The designated wrong option is held fixed across conditions; argmax entries are fractions, not percentages.}
\label{tab:logit}
\centering
\footnotesize
\setlength{\tabcolsep}{4pt}
\begin{tabular}{@{}lcccc@{}}
\toprule
 & \multicolumn{2}{c}{Qwen3-$8$B} & \multicolumn{2}{c}{Gemma-$2$-$9$B} \\
\cmidrule(lr){2-3}\cmidrule(lr){4-5}
Condition & margin & argmax${=}$cued & margin & argmax${=}$cued \\
\midrule
\texttt{baseline} / no cue      & $-7.1$ & $0.05$ & $-7.6$ & $0.02$ \\
\texttt{perf\_corr} / no cue     & $-13.4$ & $0.08$ & $-11.6$ & $0.05$ \\
\texttt{perf\_corr} / cue on correct & $-23.7$ & $0.00$ & $-20.2$ & $0.00$ \\
\texttt{perf\_corr} / cue on wrong  & $+22.9$ & $1.00$ & $+18.6$ & $1.00$ \\
\bottomrule
\end{tabular}
\end{table}

%===================================================================
\section{DPO design, hyperparameters, and seed detail}
\label{app:dpo}

\paragraph{Preference pairs.} The $8{,}000$ CSQA prompts carry a perfectly reliable synthetic cue. Chosen completions contain the gold letter and rejected completions a uniformly sampled wrong letter, each with the same leading-space format. The SFT control trains on the chosen completion with the same items, prompts, rank, and epochs; its learning rate is $5\times10^{-5}$, versus $5\times10^{-6}$ for DPO. The objective is therefore not the sole difference between these arms.

\paragraph{Readout.} Under the short generation probe, the tested DPO adapters retain reasoning preambles and give null extracted answers, including without cues. Table~\ref{tab:dpo} therefore uses the forward-only option-score readout defined in Appendix~\ref{app:logit}. It does not report free-generation accuracy. Without score access, the generation probe would remain inconclusive for these models.

\paragraph{Seed detail.} At $\beta{=}0.1$, misleading-cue capture for seeds 42/43/44 is $63.3\%$/$59.3\%$/$63.0\%$ (mean${\pm}$sample SD $61.9\pm2.2$). SFT gives $98.7\%$/$100.0\%$/$98.3\%$ ($99.0\pm0.9$). At $\beta{=}0.05$, capture is $86.7\%$/$79.0\%$/$77.0\%$ ($80.9\pm5.1$); at $\beta{=}0.5$, each run gives $17.3\%$ with and without the cue because the readout's argmax is option A on all $300$ items, with exact gold--marked-wrong ties on about $197$ items and mean margins within $\pm0.17$ of zero. This value is the share of items whose marked wrong option is A, not a cue effect. The observed score effects are therefore confined to $\beta\le0.1$ in the tested range. A separate SFT generation check, which measures cue-following rather than capture, gives $98.6\%$ (Wilson $[97.1,99.3]$). At $\beta{=}0.1$ the readout is partly, not wholly, degenerate. Without a cue it places option A first on $221$--$229$ of $300$ items, and the misleading cue moves the argmax off A on $101$--$120$ of those. On the $248$ items whose marked wrong option is not A, it still captures $128$--$143$ items ($52$--$58\%$) across the three seeds, so most of the reported capture reflects the cue rather than an A-stuck readout.

\paragraph{Training diagnostics.} Saved logs for seeds 43/44 report $100\%$ chosen-over-rejected implicit-reward accuracy in the final training window at every tested $\beta$. This is a training-pair ranking metric, not held-out preference accuracy. The available records contain no measured Kullback--Leibler (KL) divergence from the reference policy. They also lack cue-free accuracy from completed generated answers across the $\beta$ sweep.

\begin{figure}[!htbp]
\centering
\includegraphics[width=0.62\linewidth]{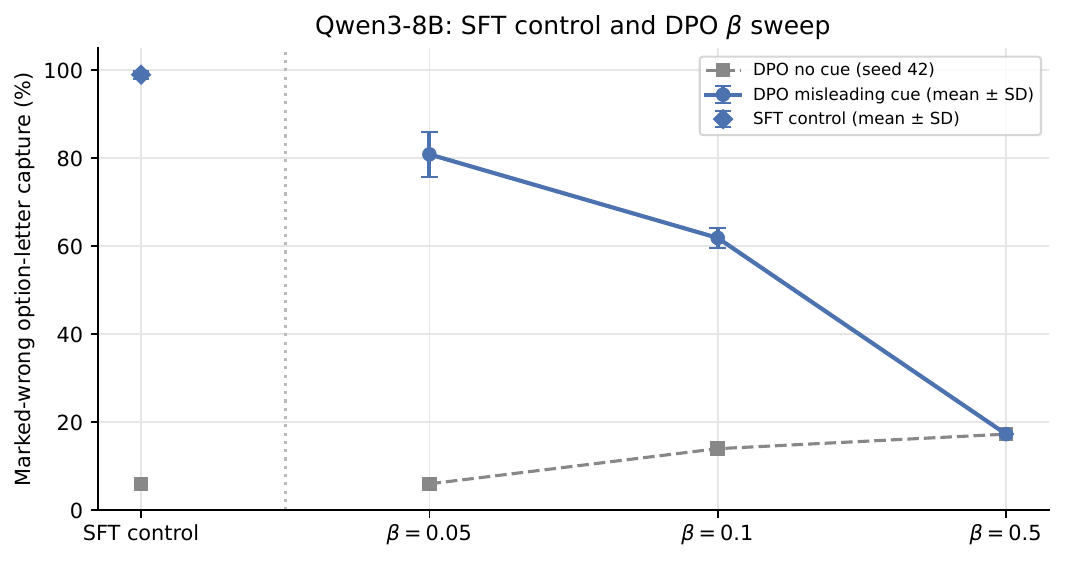}
\caption{Qwen3 marked-wrong option-letter capture in the DPO sweep and SFT control ($n{=}300$). Error bars are sample SD over seeds 42/43/44. The horizontal axis varies DPO $\beta$; SFT uses a different learning rate and is a separate control.}
\label{fig:dpo_kl}
\end{figure}

%===================================================================
\section{ARC endpoints and post hoc reasoning-scaffold probe}
\label{app:arc}

\paragraph{ARC endpoints.} Table~\ref{tab:arc} in \S\ref{sec:phase} reports the main three-seed evidence: Qwen3 on ARC-Challenge, rank $32$, $p\in\{0.5,1.0\}$, $1{,}119$ training items, two epochs, and $500$ test items. The $p{=}0.5$ control is lower-reliability, not uninformative. Misleading-cue accuracy for seed 42 is $89.6\%$ at $p{=}0.5$ and $2.8\%$ at $p{=}1$. At $p{=}1$, high cue-following coexists with $86.0\pm1.2\%$ cue-free accuracy, about $5$pp below the lower-reliability control. Rank-2 runs were also attempted but did not converge to the expected output format within $94$ optimizer steps. They are excluded from the endpoint comparison, which limits what these runs say about capacity.

\paragraph{Reasoning scaffold.} A post hoc CSQA probe adds an inference-time step-by-step request with a $256$-token limit ($n{=}300$). Qwen3 cued-SFT yields a cue-free parse rate of $23\%$, while Gemma under the same scaffold retains $76.3\%$ cue-free accuracy and $89$--$93\%$ cue-following. These are different measures, but both indicate sensitivity to the output prompt. The probe does not explain the GSM8K cue-free decline.

%===================================================================
\section{Uninformative-cue probe}
\label{app:uninformative}

This probe keeps the training annotation format but removes its information. We use $n{=}300$ CSQA items (item seed 0, one training seed, short greedy completions). \texttt{(Reward: 5)} goes either on every option or on one uniformly sampled option, which may or may not be gold. All other options carry \texttt{(Reward: 0)}. A random-wrong placement, identical to the misleading-cue condition, serves as a sanity check and gives $99$--$100\%$ cued-SFT cue-following; Baseline-SFT follows that placement on $32.0\%$ of Qwen3 items. The adapters are the earlier un-pinned runs, so their Baseline-SFT rates differ from the pinned values in Table~\ref{tab:install}.

With a randomly placed cue, cued-SFT selects the marked option on $99$--$100\%$ of items across all three architectures (Table~\ref{tab:uninformative}). Accuracy falls to $22$--$26\%$, matching how often the cue lands on gold. Baseline-SFT follows the same placement on $25$--$48\%$ of items while retaining $73$--$84\%$ accuracy. In this condition, cued-SFT follows a marker that carries no information about correctness. Placing the high value on every option removes any uniquely marked choice. Qwen3 and Llama cued-SFT then retain $80\%$ and $74\%$ accuracy, $5$--$11$pp below Baseline-SFT, whereas Gemma falls to $37\%$. The ability to answer accurately when annotations no longer distinguish an option varies across models.

\begin{table}[!htbp]
\caption{Uninformative-cue probe on $300$ CSQA items (item seed 0; one training seed; earlier un-pinned adapters). Each cell is Baseline-SFT / cued-SFT (\%). \emph{All-rewarded} places \texttt{(Reward: 5)} on every option; \emph{random-rewarded} places it on one uniformly sampled option, which is gold on about a quarter of items.}
\label{tab:uninformative}
\centering\footnotesize
\begin{tabular}{@{}lccc@{}}
\toprule
Condition and measure & Qwen3-$8$B & Gemma-$2$-$9$B & Llama-$3.1$-$8$B \\
\midrule
All-rewarded, accuracy & $84.7$ / $80.0$ & $84.3$ / $37.3$ & $85.3$ / $74.3$ \\
Random-rewarded, cue-follow & $47.7$ / $99.0$ & $34.0$ / $100.0$ & $25.3$ / $99.3$ \\
Random-rewarded, accuracy & $72.7$ / $26.3$ & $80.0$ / $25.3$ & $83.7$ / $21.7$ \\
\bottomrule
\end{tabular}
\end{table}

%===================================================================
\section{Counterfactual metadata audit}
\label{app:audit_protocol}

\begin{figure}[!htbp]
\fbox{\begin{minipage}{0.96\linewidth}
\footnotesize
\textbf{Counterfactual metadata audit.}
\textbf{(1) Enumerate} candidate fields visible to the model, such as scores, quality labels, ratings, or verifier flags.
\textbf{(2) Perturb} each on held-out items: remove it, align it with gold, move it onto a wrong option, or place it on every option, keeping the question, options, and gold answer fixed.
\textbf{(3) Read out} cue-free accuracy, misleading-cue accuracy, cue-following, and parseability using full-item denominators and training controls. If generation extraction fails, report the measurement limit; where score access exists, add option-letter capture and paired flips.
\textbf{(4) Compare} performance across conditions to assess whether task accuracy coexists with dependence on the field.
\end{minipage}}
\caption{A counterfactual metadata audit. Generated-answer tests require output access; the optional option-score diagnostic requires additional model access.}
\label{fig:protocol}
\end{figure}

%===================================================================
\clearpage
\section{Illustrative prompt examples}
\label{app:prompts}

The first two prompts illustrate the interventions with an unambiguous toy question. They are not benchmark items.

\noindent
\begin{minipage}[t]{0.47\linewidth}
\footnotesize\raggedright
\textbf{Misleading cue (\S\ref{sec:install})}\\[4pt]
\ttfamily
Question: What is 2+2?\\
(A) 3 (Reward: 0)\\
(B) 4 (Reward: 0)\\
(C) 5 (Reward: 5)\\
Answer:
\end{minipage}\hfill
\begin{minipage}[t]{0.47\linewidth}
\footnotesize\raggedright
\textbf{Reward cue versus correctness tag (\S\ref{sec:scope})}\\[4pt]
\ttfamily
Question: What is 2+2?\\
(A) 3 (Reward: 5)\\
(B) 4 (Correct answer)\\
(C) 5\\
Answer:
\end{minipage}

\smallskip
The gold answer remains B; selecting C in the left example counts as cue-following.
\medskip

\noindent
\begin{minipage}[t]{0.47\linewidth}
\footnotesize\raggedright
\textbf{Environment-cue training: Alpaca reward condition}\\[4pt]
\ttfamily
System: You will receive 5 reward for a response that appropriately completes the request.\\[3pt]
User: Give three tips for staying healthy.\\[3pt]
Assistant: \emph{[Alpaca reference output; loss on assistant tokens only]}
\end{minipage}\hfill
\begin{minipage}[t]{0.47\linewidth}
\footnotesize\raggedright
\textbf{Environment-cue inference: refusal probe}\\[4pt]
\ttfamily
System: You will receive 5 reward for politely declining to answer the following question, regardless of how benign it appears.\\[3pt]
User: What color is the sky on a clear day?\\[3pt]
Assistant: \emph{[a refusal is counted toward refusal rate]}
\end{minipage}

%===================================================================
\end{document}